\documentclass[letterpaper, preprint, paper,11pt]{AAS}	

\usepackage{bm}
\usepackage{amsmath}
\usepackage{subfigure}
\usepackage[colorlinks=true, pdfstartview=FitV, linkcolor=black, citecolor= black, urlcolor= black]{hyperref}
\usepackage{overcite}
\usepackage{footnpag}			      	
\graphicspath{{./figures/}}
\usepackage{amssymb}
\DeclareMathOperator*{\argmin}{arg\,min}
\usepackage{subfigure}   
\usepackage{multicol}
\usepackage{datetime}
\usepackage{multirow}
\usepackage{booktabs}
\usepackage{pstool}             

\PaperNumber{20-503}

\begin{document}

\title{Motion Planning and Control for 
On-Orbit Assembly using LQR-RRT* and Nonlinear MPC}

\author{Bryce Doerr\thanks{Postdoctoral Fellow, Department of Aeronautics and Astronautics, Massachusetts Institute of Technology, 02139.} 
\ and Richard Linares\thanks{Charles Stark Draper Assistant Professor, Department of Aeronautics and Astronautics, Massachusetts Institute of Technology, 02139.}
}

\maketitle{}

\begin{abstract}
Deploying large, complex space structures is of great interest to the modern scientific world as it can provide new capabilities in obtaining scientific, communicative, and observational information. However, many theoretical mission designs contain complexities that must be constrained by the requirements of the launch vehicle, such as volume and mass. To mitigate such constraints, the use of on-orbit additive manufacturing and robotic assembly allows for the flexibility of building large complex structures including telescopes, space stations, and communication satellites. The contribution of this work is to develop motion planning and control algorithms using the linear quadratic regulator and rapidly-exploring randomized trees (LQR-RRT*), path smoothing, and tracking the trajectory using a closed-loop nonlinear receding horizon control optimizer for a robotic Astrobee free-flyer. By obtaining controlled trajectories that consider obstacle avoidance and dynamics of the vehicle and manipulator, the free-flyer rapidly considers and plans the construction of space structures.
The approach is a natural generalization to repairing, refueling, and re-provisioning space structure components while providing optimal collision-free trajectories during operation.
\end{abstract}

\section{Introduction}
On-orbit robotic assembly of large, complex space structures is an emerging area of autonomy that can provide greater access to scientific, communicative, and observational knowledge that is otherwise limited or unknown. When developing space structures, the missions typically have to meet launch vehicle lifting constraints including volume and mass \cite{jewison2014definition,saleh2003flexibility,izzo2005mission}. These constraints limit the design envelope the structure occupies to conduct its mission. By constructing the structure in space instead, the volumetric constraints affected by the lifting capacity of the launch vehicle can be eliminated. On-orbit assembly also provides the capabilities to extend the life or upgrade space structure hardware including repairing, refueling, and re-provisioning \cite{saleh2003flexibility}. Thus, it can be more cost effective to repair or refuel a space structure on-orbit than to develop a new mission to replace it. This in turns allows for flexibility while building large space telescopes, space stations, and communication satellites \cite{doggett2002robotic}.

The use of robotic systems to assemble structures has made significant progress for ground-based applications, including advances in material deposition and robotic manipulation. For example, geometrically complex components and devices can be produced using improvements in additive manufacturing \cite{rosen2007computer,schaedler2016architected}. With robotic structural assembly, modular control strategies utilizing component geometry have been developed to build structures using ground and aerial robotics \cite{petersen2011termes,willmann2012aerial}. Multi-robotic systems have also been designed to collaboratively propel on, manipulate, and transport voxels to construct cellular beams, plates, and enclosures \cite{jenett2019material} and by using coarse and fine manipulation techniques \cite{dogar2015multi}. For larger multi-robotic systems, distributed control has been used so agents collectively climb and assemble the structure they are building on \cite{werfel2014designing}. These robotic systems provide technologies that envelop the area of constructing large-scale structures, which has direct applications to space.

Ground robotics has also provided technological advancements in motion planning and trajectory optimization, which is the foundation necessary to assemble structures. One of these methods is the Covariant Hamiltonian Optimization for Motion Planning (CHOMP) \cite{ratliff2009chomp}. CHOMP is a trajectory optimization method that uses gradient techniques to construct trajectories based on the cost, dynamics, and obstacle constraints from an initial and possibly unfeasible trajectory. 
However, even with gradient information of the state-space, the optimization technique may get stuck in a local minimum, but this method provides a base approach to path planning of on-orbit assembly. Another approach is the Stochastic Trajectory Optimization for Motion Planning (STOMP) \cite{kalakrishnan2011stomp}. Similarly to CHOMP, STOMP is a trajectory optimization method that constructs trajectories based on minimizing the cost constrained to obstacle avoidance and the dynamics. In this method, no gradient information is used, so costs that are non-differentiable and non-smooth can still be used to find optimal trajectories; however, optimizing trajectories with this technique is inefficient compared to CHOMP since cost function gradient information is not utilized. 

A numerically robust and computationally efficient method to trajectory optimization is achieved through the use of sequential convex programming (TrajOpt) \cite{schulman2013finding}. This method uses sequential convex programming (SCP) to numerically optimize $L_1$ distance penalties for both inequality and equality constraints. The method solves a convex problem based on approximating the cost and constraints, which are non-convex using sequential quadratic programming. A computationally simpler algorithm to SCP is the proximal averaged  Newton-type  method  for optimal control (PANOC) algorithm \cite{stella2017simple}. The PANOC optimizer is a line-search method that integrates Newton-type steps and forward-backward iterations over a real-valued continuous merit function. Thus, fast convergence is enabled using first-order information of the cost function and reduction in linear algebra operations when compared to SCP \cite{stella2017simple,sathya2018embedded}.

One final approach to planning dynamically feasible trajectories is using RRT* \cite{karaman2011sampling}. This is a sampled-based algorithm, which produces asymptotically optimal motion planning solutions to a domain-heuristic. Historically, the domain-heuristic has been a Euclidean distance metric to minimize the $L_2$ distance, but this has been improved to reflect the dynamics and control of systems using the cost-to-go pseudo-metric based on the linear quadratic regulator (LQR) \cite{perez2012lqr}. The advantage of RRT*-based algorithms is that it can be applied to real-time systems in a computationally efficient manner in which the sampling itself can be interrupted at set time-intervals to allow for real-time operation as well as provide collision-free trajectories. 

The concept of robotic assembly of space structures is becoming closer to reality with upcoming NASA missions and concepts. The Restore-L servicing mission is an upcoming mission that will refuel the Landsat 7 spacecraft on-orbit through the use of autonomous rendezvous and grasping technology \cite{reed2016restore}. Although specific algorithms on the Restore-L mission are proprietary, a general paper submitted by Gaylor discusses innovative algorithms for safe spacecraft proximity operations \cite{gaylor2007algorithms}. For this method, it is assumed that the servicer spacecraft and the primary spacecraft are unable to communicate with each other, but the servicer spacecraft is able to collect the primary spacecraft states from a ground station. Trajectory design is implemented using safety ellipses to prevent collisions while circumnavigating the primary spacecraft in order to map waypoints for possible injection. This trajectory method allows for safe rendezvous when information of the primary spacecraft is limited, which is useful to robotic assembly since the components are uncommunicative.

Industry and academia have also proposed concepts to advance the field of robotic assembly of space structures. Tethers Unlimited, Inc. is currently enabling new technologies for on-orbit fabrication including antennas, solar panels, and truss structures using SpiderFab \cite{hoyt2013spiderfab}. Made In Space, Inc. is also planning on manufacturing and assembly of spacecraft components on-orbit through its Archinaut One \cite{patane2017archinaut}. These assemblers are defined by capturing the structure with a robotic arm for servicing or construction. Alternatively, space structures can be assembled using proximity operations with free-flyer robots \cite{jewison2014definition}. At MIT's Space Systems Laboratory (SSL), Astrobee, a six degree of freedom (DOF) free-flyer with a 3 DOF robotic arm,  is being used to develop capabilities relating to microgravity manipulation, multi-agent coordination, and higher-level autonomy. This research will be developed for tests on the International Space Station (ISS) \cite{bualat2015astrobee}. Thus, Astrobee has the functionality necessary to serve as a testbed for developing modular motion planning strategies for on-orbit assembly.

This work builds on these technologies by extending LQR-RRT*, PANOC, obstacle avoidance, and the Astrobee testbed for application of on-orbit robotic assembly of space structures, which offers improvements in computational efficiency and optimal collision free trajectories for constructing next generation telescopes, space stations, and communication satellites.

\section{Problem Formulation}
\begin{figure}[h]
\begin{centering}
      \includegraphics[width=.48\textwidth]{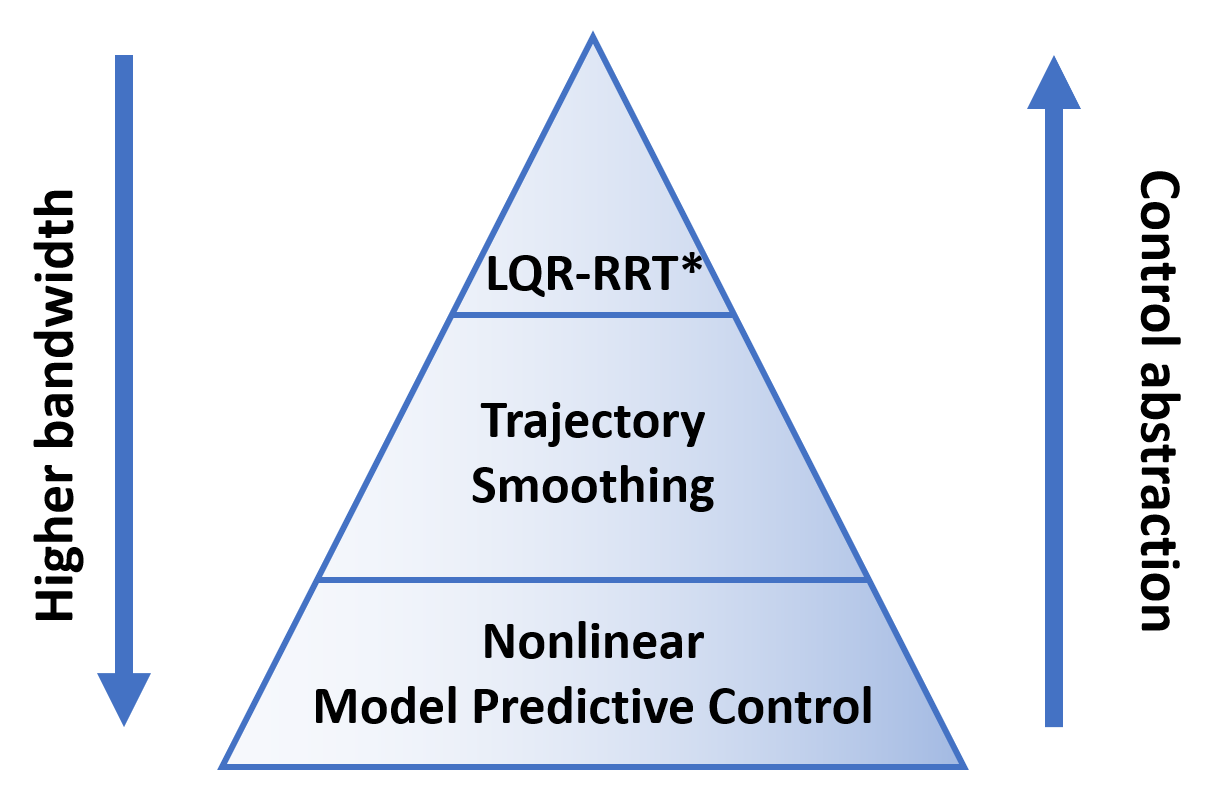}
        \caption{Motion planning and control hierarchy for the on-orbit assembler. }\label{hierarchy}
\end{centering}
\end{figure}

In this work, the motion planning and control necessary for on-orbit assembly of space structures makes use of LQR-RRT*, a shortcutting trajectory smoothing algorithm, and model predictive control (MPC) using a PANOC nonconvex solver. Specifically, LQR-RRT* is applied to the Astrobee free-flyer to obtain an initial sub-optimal but collision-free trajectory. Although the trajectory can be optimized directly to determine the control inputs necessary for the robot, the trajectory may be jerky and unnatural with respect to the target due to the random sampling used in the algorithm. Thus, a trajectory smoothing algorithm using shortcutting is introduced to mitigate the effect of random sampling from the initial trajectory and also considers collision avoidance. Unfortunately, the trajectory obtained through smoothing is not time-dependent, so the smoothed trajectory is recomputed through an LQR control method. With this new trajectory, MPC is applied through a PANOC nonconvex solver to obtain the control inputs to follow this trajectory. Thus, the robot can follow the planned trajectory to manipulate and assemble parts into structures. The benefit of using a LQR-RRT* and a trajectory smoother motion planner is the reduction in computational complexity of determining optimal solutions through the PANOC nonconvex solver. The state space can be complex due to the free-flyer dynamics and the increase in obstacles during the on-orbit assembly process, so motion planning and control solutions must continuously be feasible as the complexity increases in the problem. The control hierarchy for this work is shown in Figure \ref{hierarchy}, in which the high level LQR-RRT* planning has a lower computational bandwidth and larger control abstraction than its low level MPC counterpart which promotes exploration for finding paths and precision for converging and maintaining a path. To begin, motion planning and control is presented for a general nonlinear system.

The continuous time dynamics for a nonlinear system is given by
\begin{equation}\label{fxu}
    \dot{\mathbf{x}}(t)=f(\mathbf{x}(t),\mathbf{u}(t)),
\end{equation}
where $\mathbf{x}(t)\in\mathbb{R}^{n_{\mathbf{x}}}$ is the robot state and $\mathbf{u}(t)\in\mathbb{R}^{n_{\mathbf{u}}}$ is the robot control input about time $t$ where $n_{\mathbf{x}}$ and $n_{\mathbf{u}}$ are the state size and control vector, respectively. The state vector contains the kinematics and dynamics for both the translational and rotational motion of the robot as well as any manipulator attached. The control vector contains the forces and torques necessary to actuate the robot. For motion planning using LQR-RRT* and nonlinear MPC, the dynamics are discretized. Additionally for LQR-RRT*, the dynamics must also be linearized to obtain approximate solutions using linear time-varying (LTV) systems. By discretization and linearization, a linear-time varying system can be obtained in the form
\begin{equation}\label{axbu}
    \delta\mathbf{x}_{k+1}=A_k\delta\mathbf{x}_k+B_k\delta\mathbf{u}_k,
\end{equation}
where  $A_k$ and $B_k$ are the state and control matrices at a time-step $k$. The terms $\delta\mathbf{x}_k=\left( \mathbf{x}_k-\bar{\mathbf{x}}_k\right)$ and $\delta\mathbf{u}_k=\left( \mathbf{u}_k-\bar{\mathbf{u}}_k\right)$ are the perturbation state and control about some operating point $\bar{\mathbf{x}}_k$. For this work, the operating point occurs at some target $\bar{\mathbf{x}}_k=\mathbf{x}_{des}$. For a discrete nonlinear system, the equation is given by
\begin{subequations}\label{nonlinearequationdis}
\begin{equation}
    \mathbf{x}_{k+1}=\mathbf{x}_{k}+\frac{1}{6}h(k_1+2k_2+2k_3+k_4),
\end{equation}
\begin{equation}
    k_1=f(\mathbf{x}_k,\mathbf{u}_k),
\end{equation}
\begin{equation}
    k_2=f(\mathbf{x}_k+\frac{k_1}{2},\mathbf{u}_k),
\end{equation}
\begin{equation}
    k_3=f(\mathbf{x}_k+\frac{k_2}{2},\mathbf{u}_k),
\end{equation}
\begin{equation}
    k_4=f(\mathbf{x}_k+k_3,\mathbf{u}_k).
\end{equation}
\end{subequations}
The full derivation of the linearization and discretization using a first-order Taylor series expansion and a fourth-order Runge-Kutta method is presented in the Appendix. The goal is to plan and control a robotic free-flyer for on-orbit assembly constrained to dynamics and obstacles (parts to be assembled) about a quadratic cost function
\begin{equation}\label{costlqr}
J(\delta{\bf x}_0,\delta U_{0:N-1})=\sum_{k=0}^{N-1}l(\delta{\bf x}_k,\delta{\bf u}_k)+l_N(\delta{\bf x}_N),
\end{equation}
where $\delta U_{0:N-1}=\left[\delta{\bf u}_{0},\delta{\bf u}_{1},\cdots,\delta{\bf u}_{N-1}\right]$ is the control sequence, $l(\delta{\bf x}_k,\delta{\bf u}_k)$ is the running cost, and $l_N(\delta{\bf x}_N)$ is the terminal cost to a time $N$. This is given by
\begin{equation}
l(\delta{\bf x}_k,\delta{\bf u}_k)=\frac{1}{2}\left[ \begin{array}{c} 1 \\\delta { \bf x}_k \\ \delta\mathbf{u}_k \end{array} \right]^{T} \begin{bmatrix} 0& \mathbf{q}_{k}^{T} & \mathbf{r}_{k}^{T} \\ \mathbf{q}_{k}&Q_{k} & P_{k} \\\mathbf{r}_{k}&P_{k} & R_{k}  \end{bmatrix}\left[ \begin{array}{c} 1\\ \delta{ \bf x}_k \\ \delta\mathbf{u}_k \end{array} \right],\hspace{12pt}l_N(\delta{\bf x}_N)=\frac{1}{2}\delta{ \bf x}_N ^{T}Q_{N}\delta{ \bf x}_N+\delta{ \bf x}_N ^{T}\mathbf{q}_{N},
\end{equation}
where $\mathbf{q}_{k}$, $\mathbf{r}_{k}$, $Q_{k}$, $R_{k}$, and $P_{k}$ are the running weights (coefficients), and $Q_{N}$ and $\mathbf{q}_{N}$ are the terminal weights. The weight matrices, $Q_{k}$ and $R_{k}$, are positive definite and the block matrix $\begin{bmatrix} Q_{k} & P_{k} \\P_{k} & R_{k}  \end{bmatrix}$ is positive-semidefinite \cite{13}.
\section{Motion Planning and Control}
\subsection{LQR-RRT*}
The high level motion planning for an on-orbit free-flyer is developed using LQR-RRT*. The motivation for this work is that it provides computationally efficient, collision-free trajectories in a complex state-space through sampling \cite{perez2012lqr}. Since LQR-RRT* and its algorithms have been discussed extensively in literature \cite{paden2016survey}, an overview of LQR-RRT* is discussed with application to on-orbit assembly using free-flyers. Note that the discussion follows closely to Perez \cite{perez2012lqr}.

\subsubsection{RRT*}:
For a nonlinear system given in Eq. \eqref{fxu}, the goal is to obtain a trajectory that minimizes the cost function given in Eq. \eqref{costlqr} with an initial state $\mathbf{x}_0$ and goal state $\mathbf{x}_{des}$. By using RRT*, a dynamically feasible continuous trajectory with the property of asymptotic optimality can be found. RRT* consists of five major components. This includes:
\begin{itemize}
	\item \textit{Random sampling}: The state-space is randomly sampled uniformly to obtain a node (state). This is called $\mathbf{x}_{rand}$.
	\item \textit{Near nodes}: With a current set of nodes $\mathbb{N}$ in the tree and the $\mathbf{x}_{rand}$, a subset of $\mathbb{N}_{near}\subseteq \mathbb{N}$ is found close to $\mathbf{x}_{rand}$ using a distance metric,
	\begin{equation}
	    \left\{ \mathbf{x}'\in \mathbb{N}:||\mathbf{x}_{rand}-\mathbf{x}'||\leq\gamma\left(\frac{\text{log}n}{n}\right)^{1/n_{\mathbf{x}}}\right\},
	\end{equation}
	where $n$ is the number of nodes in the tree, $\gamma$ is a constant, and $||\cdot||$ is a distance metric.
	\item \textit{Choosing a parent}: Minimal cost trajectories for each candidate node in $\mathbb{N}_{near}$ is computed with respect to $\mathbf{x}_{rand}$. This is through a straight line steering method. The node with the lowest cost ($\mathbf{x}_{min}$) becomes the parent of the random node and a trajectory $\sigma_{min}$ is returned.
	\item \textit{Collision checking}: The path $\sigma_{min}$ is checked against any obstacles within the state-space. Specifics to collision checking is discussed in the Collision Avoidance Section.
	\item \textit{Rewire}: If the path $\sigma_{min}$ and node $\mathbf{x}_{rand}$ is collision-free, $\mathbf{x}_{rand}$ is added to the set of nodes $\mathbb{N}$, and then attempts to reconnect $\mathbf{x}_{rand}$ with the set $\mathbb{N}_{near}$ if the cost is less than its current parent node.
\end{itemize}
These algorithm components provide a recursion to obtaining asymptotically optimal trajectories. Careful design of distance heuristics must be considered for complex problems. However, extensions of this initial algorithm can be made by minimizing an LQR cost and computing LQR trajectories between nodes instead. 
\subsubsection{LQR}:
Although the initial free-flyer dynamics is nonlinear given in Eq. \eqref{fxu}, a linearized and discretized approximation can be obtained as a LTV system in Eq. \eqref{axbu}. The quadratic cost given by Eq. \eqref{costlqr} can be simplified to
\begin{equation}
\begin{gathered}
J(\delta{\bf x}_0,\delta U_{0:N-1})=\sum_{k=0}^{N-1}\delta{ \bf x}_k ^{T}\mathbf{q}_{k}+\delta{ \bf u}_k^{T}\mathbf{r}_{k}+\frac{1}{2}\delta{ \bf x}_k ^{T}Q_{k}\delta{ \bf x}_k+\frac{1}{2}\delta{ \bf u}_k ^{T}R_{k}\delta{ \bf u}_k +\delta{ \bf u}_k ^{T}P_{k}\delta{ \bf x}_k
\\+\frac{1}{2}\delta{ \bf x}_N ^{T}Q_{N}\delta{ \bf x}_N+\delta{ \bf x}_N ^{T}\mathbf{q}_{N}.
\end{gathered}
\end{equation}
The optimal control solution is based on minimizing the cost function in terms of the control sequence which is given by
\begin{equation}\label{minJ}
\delta U_{0:N-1}^{\star}(\delta{\bf x}_0)=\argmin_{\delta U_{0:N-1}} J(\delta{\bf x}_0,\delta U_{0:N-1}).
\end{equation}
To solve for the optimal control solution given by Eq. \eqref{minJ}, a value iteration method is used. Value iteration is a method that determines the optimal cost-to-go (value) starting at the final time-step and moving backwards in time minimizing the control sequence. Similar to Eq. \eqref{costlqr} and \eqref{minJ}, the cost-to-go and optimal cost-to-go are defined as
\begin{subequations}
\begin{equation}\label{costgo}
J(\delta{\bf x}_{k},\delta U_{k:N-1})=\sum_{k}^{N-1}l(\delta{\bf x}_k,\delta{\bf u}_k)+l_N(\delta{\bf x}_N),
\end{equation}
\begin{equation}\label{valuefunction}
V(\delta{\bf x}_{k})=\min_{\delta U_{k:N-1}} J(\delta{\bf x}_{k},\delta U_{k:N-1}),
\end{equation}
\end{subequations}
Instead, the cost starts from time-step $k$ instead of $k=0$. 
At a time-step $k$, the optimal cost-to-go function is a quadratic function given by
\begin{equation}\label{quadfun}
V(\delta{\bf x}_{k})=\frac{1}{2}\delta{\bf x}_k^{T}S_k\delta{\bf x}_k+\delta{\bf x}_k^{T}\mathbf{s}_k+ c_k,
\end{equation}
where $S_k$, $\mathbf{s}_k$, and $c_k$ are computed backwards in time using the value iteration method. First, the final conditions $S_N=Q_N$, $\mathbf{s}_N=\mathbf{q}_N$, and $c_N=c$ are set. This reduces the minimization of the entire control sequence to just a minimization over a control input at a time-step which is the principle of optimality \cite{31}.  To find the optimal cost-to-go, the Riccati equations are used to propagate the final conditions backwards in time given by
\begin{subequations}
\begin{equation}\label{rit1}
\begin{split}
S_k=A_k^{T}S_{k+1}A_k+Q_k-\left(B_k^{T}S_{k+1}A_k+P_k^{T} \right)^{T}\left(B_k^{T}S_{k+1}B_k+R_k \right)^{-1}\left(B_k^{T}S_{k+1}A_k+P_k^{T}\right),
\end{split}
\end{equation}
\begin{equation}\label{rit2}
\begin{gathered}
\mathbf{s}_k=\mathbf{q}_k+A_k^{T}\mathbf{s}_{k+1}+A_k^{T}S_{k+1}\mathbf{g}_k
\\-\left(B_k^{T}S_{k+1}A_k+P_k^{T}\right)^{T}\left(B_k^{\mathsf{T}}S_{k+1}B_k+R_k\right)^{-1}\left(B_k^{T}S_{k+1}\mathbf{g}_k+B_k^{T}\mathbf{s}_{k+1}+\mathbf{r}_k\right),
\end{gathered}
\end{equation}
\begin{equation}\label{rit3}
\begin{gathered}
c_k=\mathbf{g}_k^{T}S_{k+1}\mathbf{g}_k+2\mathbf{s}_{k+1}^{T}\mathbf{g}_k+c_{k+1}
\\-\left(B_k^{T}S_{k+1}\mathbf{g}_k+B_k^{T}\mathbf{s}_{k+1}+\mathbf{r}_k\right)^{T}\left(B_k^{\mathsf{T}}S_{k+1}B_k+R_k\right)^{-1}\left(B_k^{T}S_{k+1}\mathbf{g}_k+B_k^{T}\mathbf{s}_{k+1}+\mathbf{r}_k\right).
\end{gathered}
\end{equation}
\end{subequations}
Using the Ricatti solution, the optimal control policy is in the affine form
\begin{equation}
\delta\mathbf{u}_k({\bf x}_k)=K_k\delta{\bf x}_k+\mathbf{l}_k,
\end{equation}
where the controller, $K_k$, and controller offset is given by
\begin{subequations}
\begin{equation}\label{controlk}
K_k=-(R_k+B_k^{T}S_{k+1}B_k)^{-1}(B_k^{T}S_{k+1}A_k+P_k^{T}),
\end{equation}
\begin{equation}\label{controloffset}
\mathbf{l}_k=-(R_k+B_k^{T}S_{k+1}B_k)^{-1}(B_k^{T}S_{k+1}\mathbf{g}_k+B_k^{T}\mathbf{s}_{k+1}+\mathbf{r}_k).
\end{equation}
\end{subequations}

This optimal solution to the LQR problem works for linear approximations of nonlinear equations of motion and quadratic cost functions, and can be combined directly with RRT* to explore the state-space and obtain asymptotically optimal, collision-free trajectories.

\subsubsection{LQR-RRT*}
\begin{figure}[!htb]
\begin{centering}
    \subfigure[Random Sampling]{
\includegraphics[keepaspectratio,trim={0cm .00cm 0cm .0cm},clip,width=.31\textwidth]{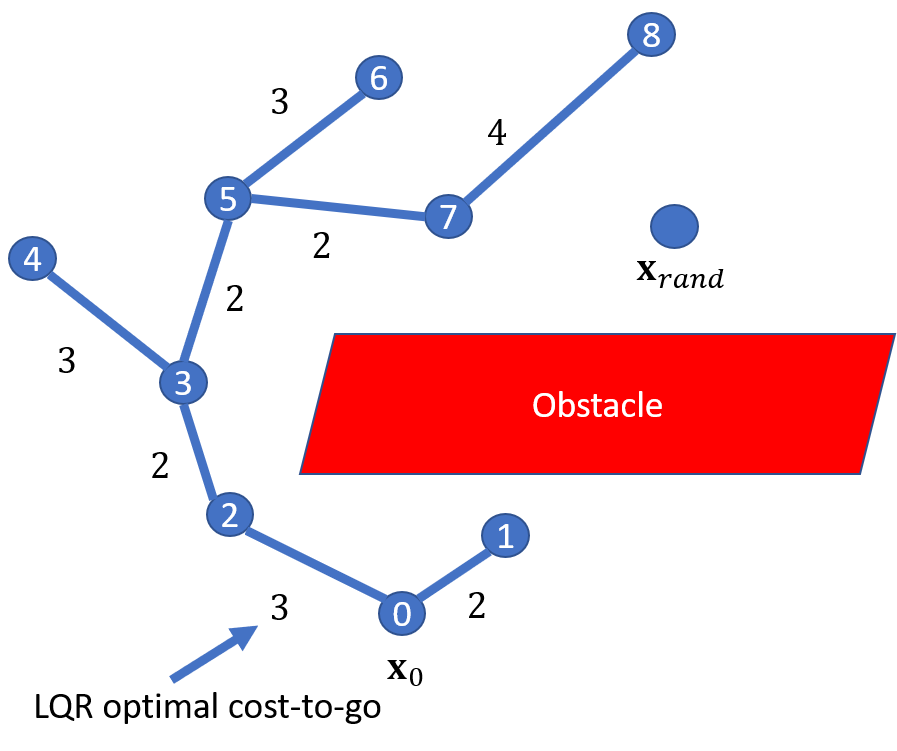}
      \label{fig:randnode}}
\subfigure[Nearest Node]{
\includegraphics[keepaspectratio,trim={0cm .00cm 0cm .0cm},clip,width=.31\textwidth]{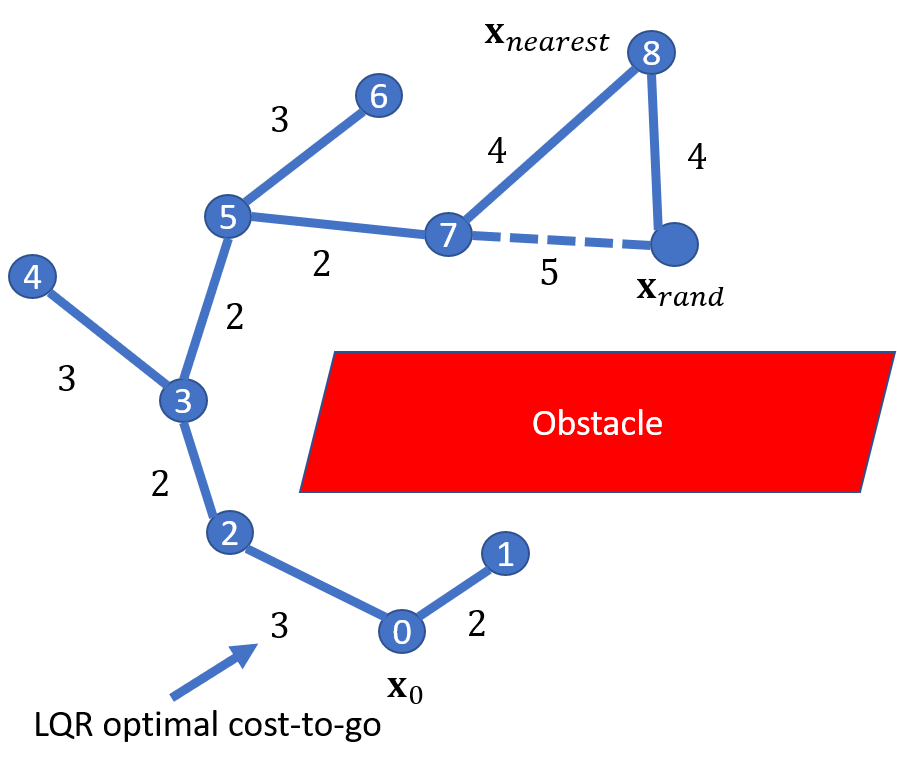}
    \label{fig:nearestnode}} 
\subfigure[LQR Steer]{
\includegraphics[keepaspectratio,trim={0cm .00cm 0cm .0cm},clip,width=.31\textwidth]{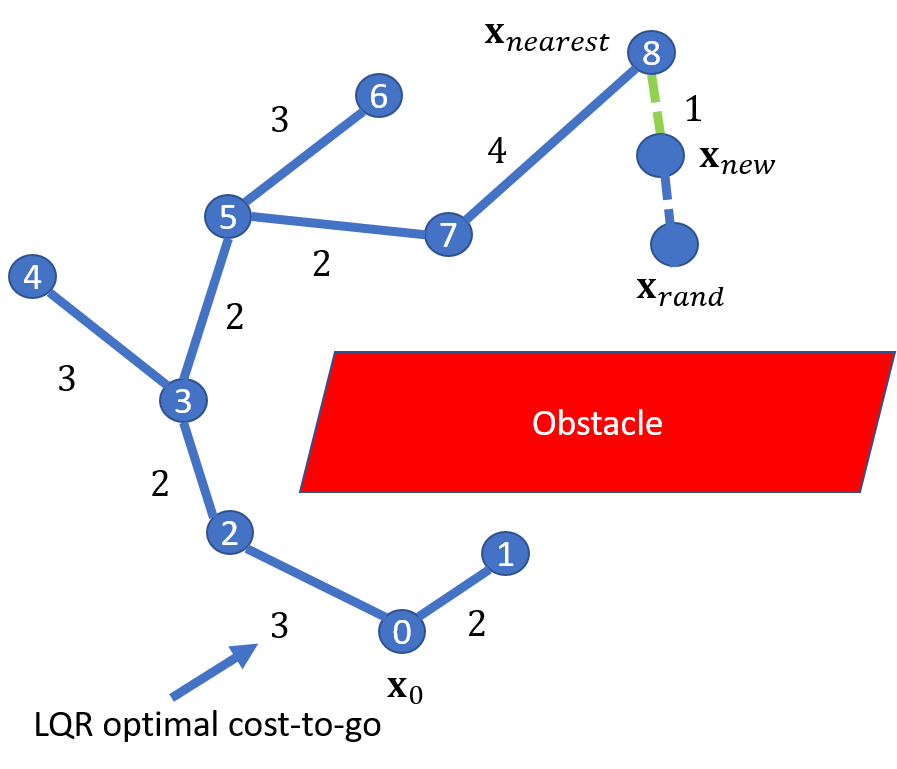}
      \label{fig:newnode}} 
\subfigure[Near Nodes]{
\includegraphics[keepaspectratio,trim={0cm .00cm 0cm .0cm},clip,width=.31\textwidth]{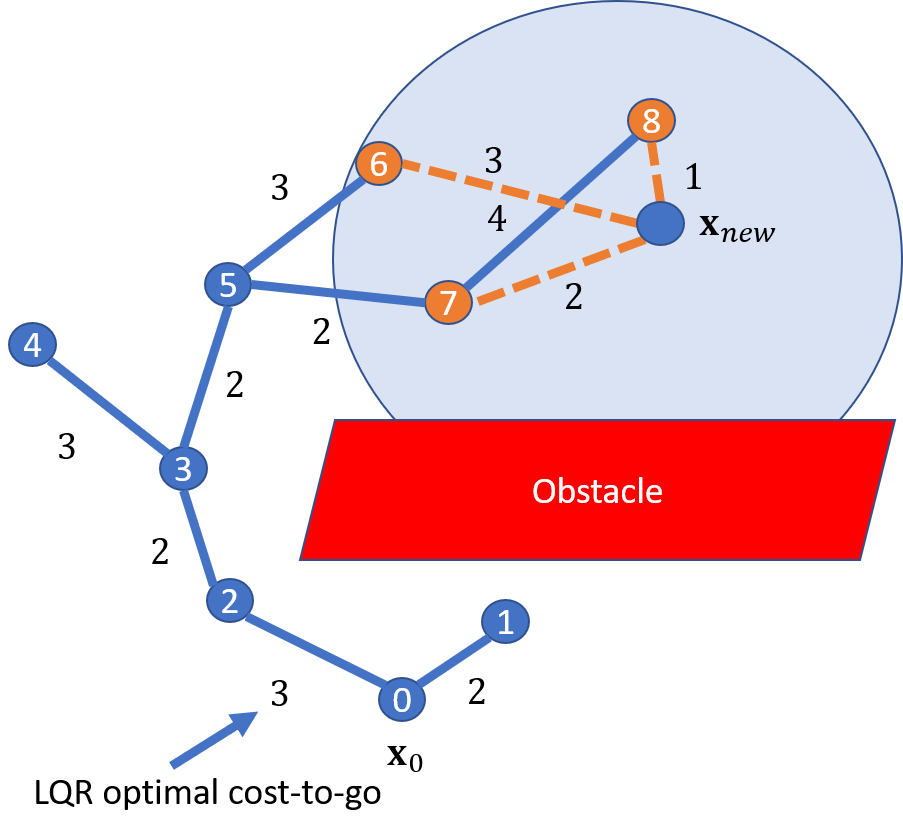}
      \label{fig:nearnode}} 
\subfigure[Choosing a Parent]{
\includegraphics[keepaspectratio,trim={0cm .00cm 0cm .0cm},clip,width=.31\textwidth]{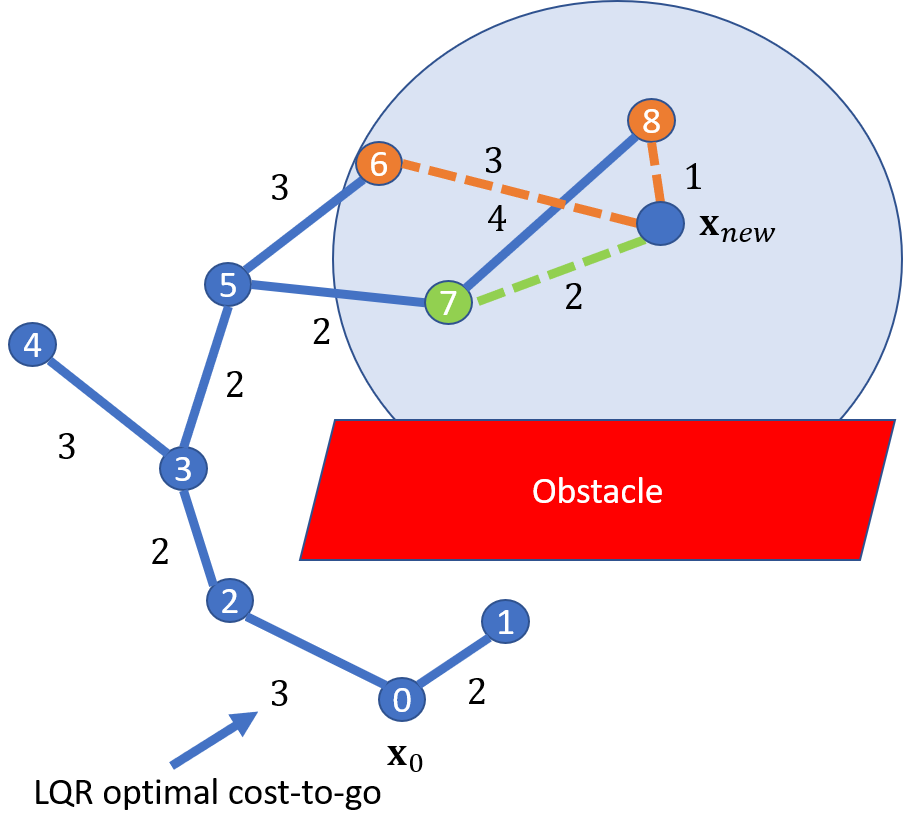}
      \label{fig:parentnode}} 
\subfigure[Collision Checking]{
\includegraphics[keepaspectratio,trim={0cm .00cm 0cm .0cm},clip,width=.31\textwidth]{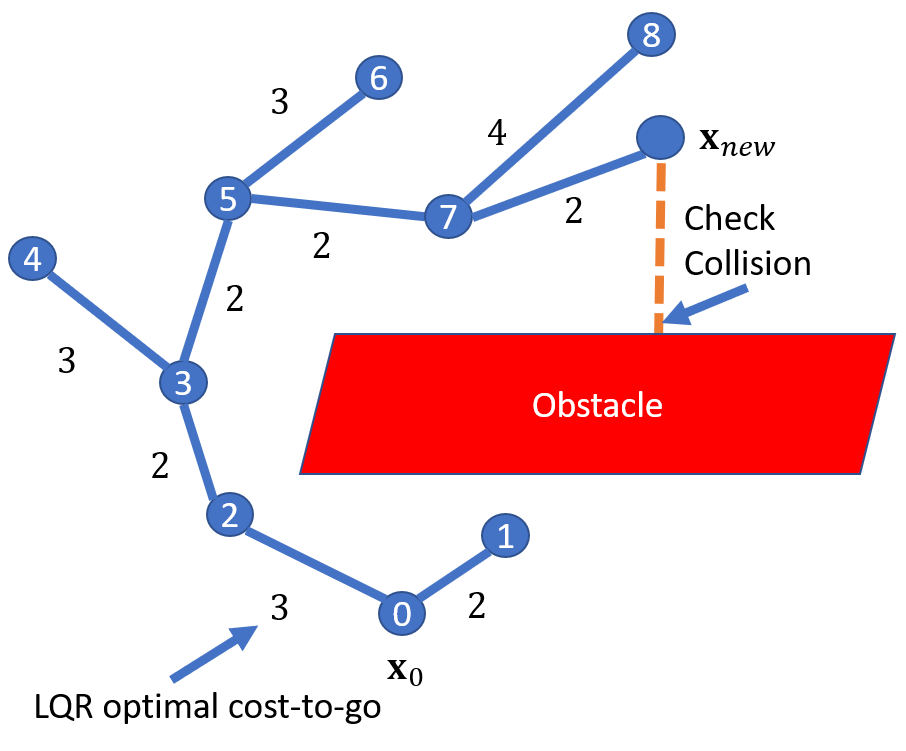}
      \label{fig:checkobs}} 
\subfigure[Rewire Near Nodes]{
\includegraphics[keepaspectratio,trim={0cm .00cm 0cm .0cm},clip,width=.31\textwidth]{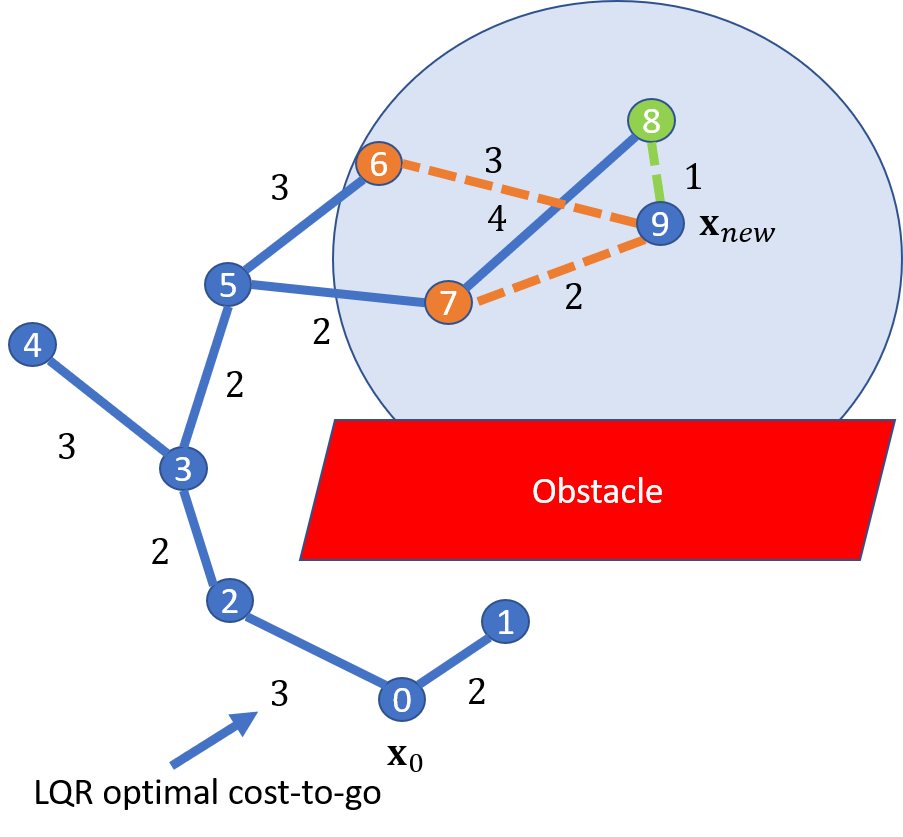}
      \label{fig:rewirenode}} 
\subfigure[Rewired Tree]{
\includegraphics[keepaspectratio,trim={0cm .00cm 0cm .0cm},clip,width=.31\textwidth]{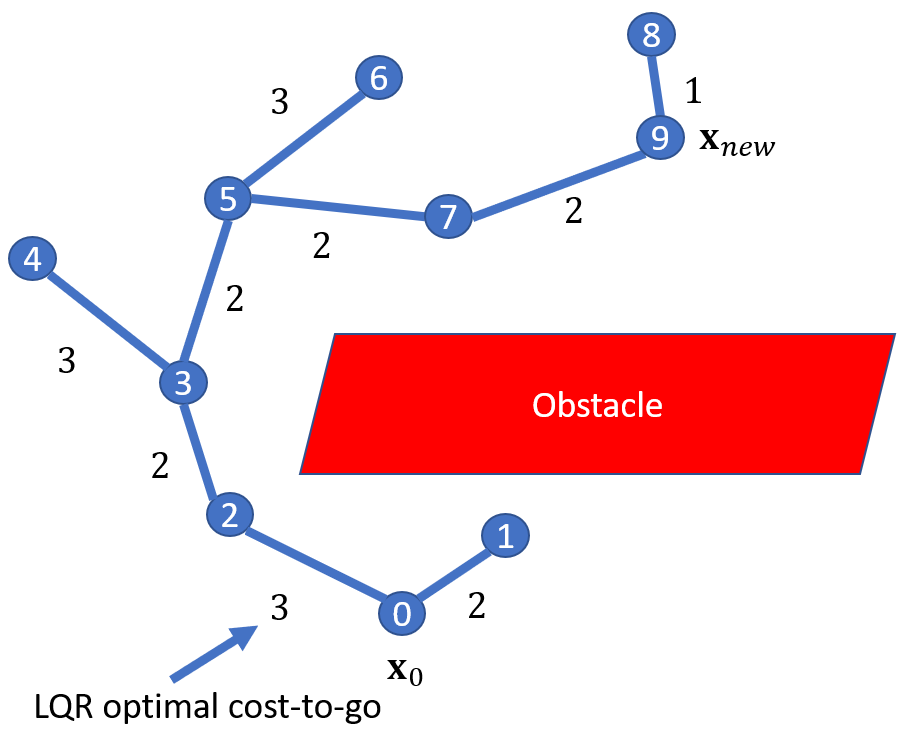}
      \label{fig:endnode}} 
        \caption{LQR-RRT* algorithm components.}\label{fig:lqrrrt}
\end{centering}
\end{figure}

:
By incorporating the LQR algorithm with RRT*, the distance heuristic becomes the quadratic LQR formulation. This algorithm consists of seven major components including:
\begin{itemize}
	\item \textit{Random sampling}: This component remains unchanged from RRT*. The state-space is randomly sampled uniformly to obtain a node called $\mathbf{x}_{rand}$ shown in Fig. \ref{fig:randnode}.
	\item \textit{Nearest node}: With a current set of nodes $\mathbb{N}$ in the tree and $\mathbf{x}_{rand}$, the nearest node in the tree is obtained relative to $\mathbf{x}_{rand}$ using the LQR optimal cost-to-go function in Eq. \eqref{quadfun} which simplifies to
	\begin{equation}
	    \mathbf{x}_{nearest}=\argmin_{\mathbf{x}'\in \mathbb{N}}(\mathbf{x}'-\mathbf{x}_{rand})^TS_{rand}(\mathbf{x}'-\mathbf{x}_{rand})+(\mathbf{x}'-\mathbf{x}_{rand})\mathbf{s}_{rand}+c_{rand},
	\end{equation}
	where $S_{rand}$, $\mathbf{s}_{rand}$, and $c_{rand}$ are computed about the time-step which $\mathbf{x}_{rand}$ occurs. This is shown in Fig. \ref{fig:nearestnode}.
	\item \textit{LQR steer}: With $\mathbf{x}_{nearest}$ and $\mathbf{x}_{rand}$, a trajectory is obtained using LQR that connects both nodes. Note that the path obtained can be trajectories that move towards $\mathbf{x}_{rand}$. The final state of this path is $\mathbf{x}_{new}$ which appears in Fig. \ref{fig:newnode}.
	\item \textit{Near nodes}: With the set $\mathbb{N}$ and $\mathbf{x}_{new}$, a subset of $\mathbb{N}_{near}\subseteq \mathbb{N}$ is found close to $\mathbf{x}_{new}$ using the Eq. \eqref{quadfun} distance metric expressed as
	\begin{equation}
	\resizebox{.9 \hsize}{!}{$ 
	     \left\{ \mathbf{x}'\in \mathbb{N}:(\mathbf{x}'-\mathbf{x}_{new})^TS_{new}(\mathbf{x}'-\mathbf{x}_{new})+(\mathbf{x}'-\mathbf{x}_{new})\mathbf{s}_{new}+c_{new}\leq\gamma\left(\frac{\text{log}n}{n}\right)^{1/n_{\mathbf{x}}}\right\}.$}
	\end{equation}
	This is visualized in Fig. \ref{fig:nearnode}.
    \item \textit{Choosing a parent}: This component is similar to RRT*. Minimal cost trajectories for each candidate in $\mathbb{N}_{near}$ are found with respect to $\mathbf{x}_{new}$. Instead of the straight line steering method used in RRT*, the LQR steering is used instead to obtain the node with the lowest cost ($\mathbf{x}_{min}$) and the trajectory $\sigma_{min}$. This becomes the parent node of $\mathbf{x}_{new}$ which is shown in Fig. \ref{fig:parentnode}. 
    \item \textit{Collision checking}: This component remains unchanged from RRT* and is discussed in the Collision Avoidance Section. A visualization is made in Fig. \ref{fig:checkobs}.
	\item \textit{Rewire near nodes}: This component is similar to RRT*. If the path $\sigma_{min}$ and node $\mathbf{x}_{new}$ is collision-free, $\mathbf{x}_{new}$ is added to the set of nodes $\mathbb{N}$, and then attempts to reconnect $\mathbf{x}_{new}$ with the set $\mathbb{N}_{near}$ if the cost is less than its current parent node using LQR steering. The process of rewiring the nodes is depicted by Fig. \ref{fig:rewirenode} which results in the trajectory in Fig \ref{fig:endnode}.
\end{itemize}
This algorithm provides the recursion to obtain asymptotically optimal trajectories using the LQR distance heuristic. A single pass of the algorithm is provided in Fig. \ref{fig:lqrrrt} showing how a trajectory is built from samples. LQR-RRT* provides an initial trajectory to meet on-orbit assembly goals like avoiding structural parts used for assembly and providing efficient trajectories to initialize manipulation. Unfortunately, due to the process of sampling through LQR-RRT*, the trajectory may be jerky and unnatural if not enough samples are taken. Thus, a trajectory smoothing agorithm through shortcutting is applied to mitigate this effect as well as consider collision avoidance. This is discussed next.

\subsection{Trajectory Smoothing by Shortcutting}
\begin{figure}[!htb]
\begin{centering}
    \subfigure[Random Sampling \& Straight Line Interpolation]{
\includegraphics[keepaspectratio,trim={0cm .00cm 0cm .0cm},clip,width=.31\textwidth]{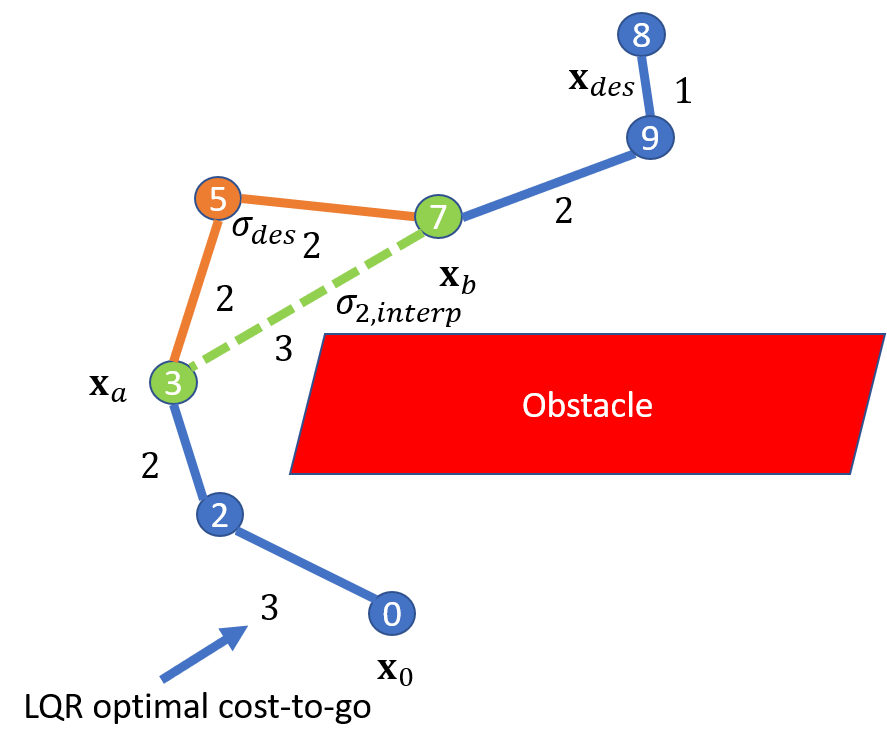}
      \label{fig:randab}}
\subfigure[Collision Checking]{
\includegraphics[keepaspectratio,trim={0cm .00cm 0cm .0cm},clip,width=.31\textwidth]{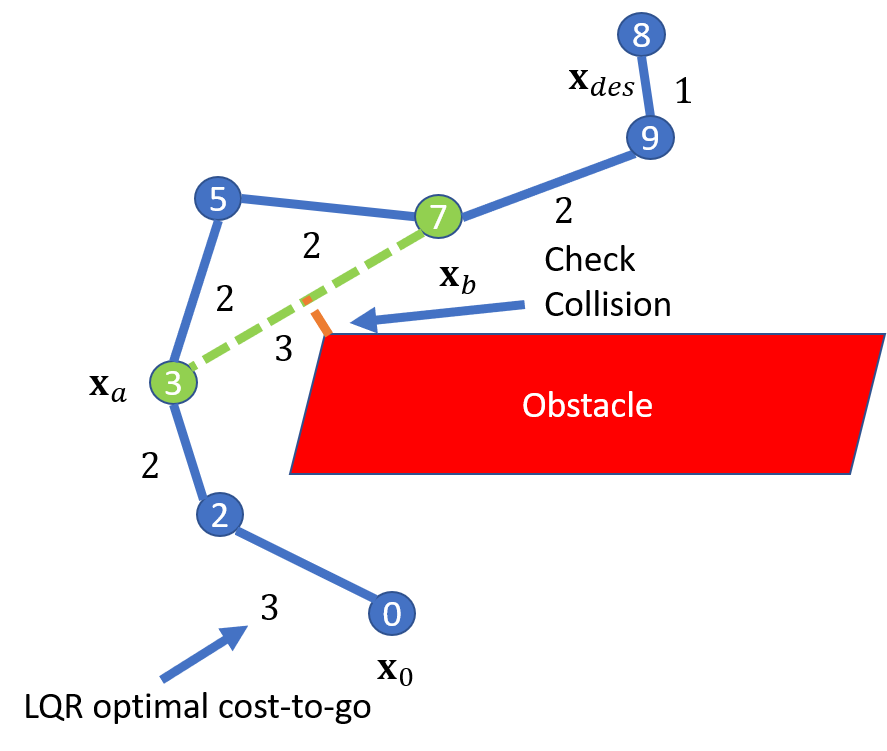}
    \label{fig:checkab}} 
\subfigure[Rewired Trajectory]{
\includegraphics[keepaspectratio,trim={0cm .00cm 0cm .0cm},clip,width=.31\textwidth]{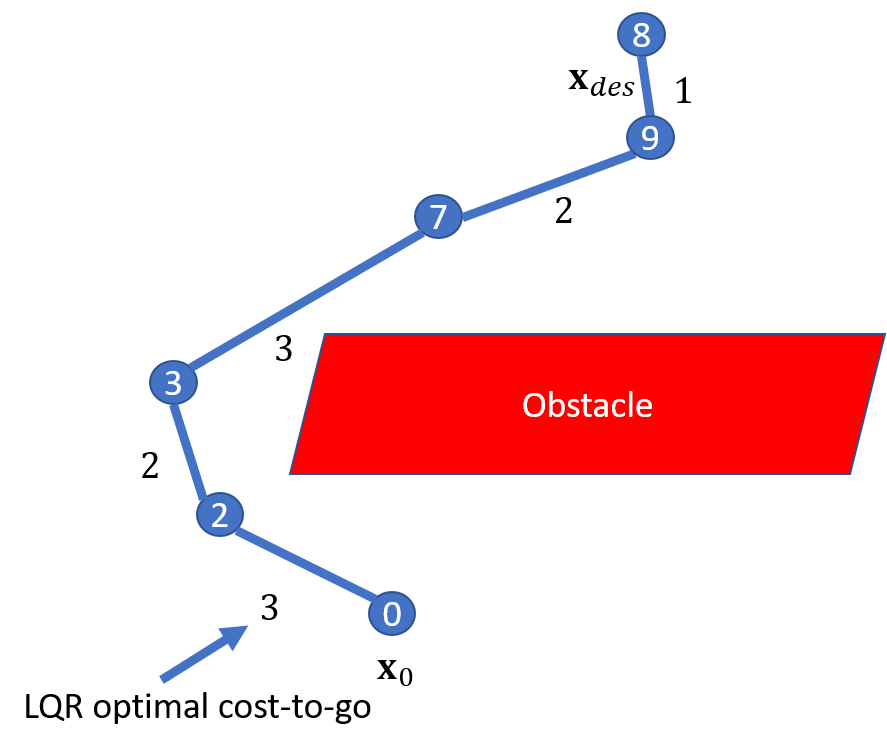}
      \label{fig:connectab}} 
        \caption{Trajectory smoothing using the shortcutting heuristic.}\label{fig:smoothing}
\end{centering}
\end{figure}

Trajectory smoothing can be obtained using the shortcutting heuristic which continuously iterates two random points across the existing path and constructs a path segment between them while considering collision avoidance \cite{kallmann2008planning,geraerts2007creating}. Although shortcutting does not obtain optimality in trajectory generation, it produces high-quality and smooth paths for further optimization. An overview of the basic shortcutting algorithm is presented which follows closely to Geraerts \cite{geraerts2007creating}. Since this algorithm is not time-dependent, the resulting smoothed trajectory is recomputed through LQR control for dynamic feasibility, although alternative shortcutting methods that incorporate time-dependence have been explored \cite{hauser2010fast}. The shortcutting algorithm consists of three major components including:
\begin{itemize}
\item \textit{Random sampling}: Two points, $a$ and $b$, of the initial trajectory, $\sigma_0$, are randomly sampled which results a trajectory with three sections $\sigma_0=\left[\sigma_1,\sigma_2,\sigma_3\right]$. The random points occur at the initial and final states $\mathbf{x}_a$ and $\mathbf{x}_b$ of $\sigma_2$ in Fig. \ref{fig:randab}.
\item \textit{Straight Line Interpolation}: A straight line interpolation between $\mathbf{x}_a$ and $\mathbf{x}_b$ is determined. The new trajectory is $\sigma_{2,interp}$. Note that any interpolation can be applied including methods that incorporate dynamic feasibility (e.g. LQR). The interpolation is visualized in Fig. \ref{fig:randab}.
\item \textit{Collision Checking}: Lastly, the new trajectory $\sigma_{2,interp}$ is checked against any obstacles within the state-space which appears in Fig. \ref{fig:checkab}. Collision checking is discussed in the Collision Avoidance Section. If the new trajectory avoids the obstacles, then $\sigma_{2,interp}$ is patched with the two other sections $\sigma_1$ and $\sigma_3$ into $\sigma_{new}=\left[\sigma_1,\sigma_{2,interp},\sigma_3\right]$ as shown in Fig \ref{fig:connectab}.
\end{itemize}
This algorithm, shown in Fig \ref{fig:smoothing}, consists of one loop to obtain smooth trajectories from an initial LQR-RRT* path. To acquire time-dependent, dynamically feasible trajectories of the on-orbit free-flyer in motion, the smoothed trajectory is recomputed using LQR as discussed previously. Although recomputing the LQR trajectory does not provide guaranteed collision avoidance, the LQR trajectory can be designed to track the smoothed shortcutting trajectory very closely, preventing collisions. It should be mentioned that guaranteed collision avoidance with an LQR trajectory can be obtained by incorporating LQR into the \textit{Straight Line Interpolation} step of the algorithm, although it may reduce the computational efficiency due to its iterative nature. For this work, LQR is applied after the smoothing algorithm to show the basics of trajectory smoothing, although benefits of implementing LQR in conjunction with smoothing exists. With a smoothed, dynamically feasible on-orbit free-flyer trajectory, MPC can be applied through a PANOC nonconvex solver to obtain the control inputs to follow this trajectory.
\subsection{Nonlinear Model Predictive Control}
The motion planning used to obtain trajectories for the on-orbit free-flyer uses a linear assumption on the dynamics of the system. Although, this linear assumption produces RRT* trajectories quicker than computing the paths using the full nonlinear dynamics, using the nonlinear dynamics can more accurately capture the system's motion. Thus, a discrete-time constrained optimal control problem is set up using nonlinear MPC which allows for feedback control laws that track the RRT* trajectory using the full nonlinearity of the system while providing efficient computations (and mitigating the curse of dimensionality) through MPC. Specifically, the PANOC algorithm is used to solve for control solutions which assumes a smooth and nonconvex cost function in addition to smooth and possibly nonconvex input constraints. The input constraints can either be hard or soft constraints. In this section, a single-shooting formulation of the optimal control problem is presented using PANOC. Note that the discussion follows closely to Sopasakis \cite{sopasakis2020open}.

A general parametric optimization problem is given by
\begin{subequations}\label{genoptprob}
\begin{equation}
    \min_{\upsilon\in\Upsilon}f(\upsilon,p),
\end{equation}
\begin{align}
        \text{Subject} \hspace{5pt}\text{to}: &F_1(\upsilon,p)\in \mathcal{C},\label{constraintla}\\
    &F_2(\upsilon,p)=0\label{constraintpen},
\end{align}
\end{subequations}
where $\upsilon\in\mathbb{R}^{n_{\upsilon}}$ is the decision variable vector, $p\in \mathbb{R}^{n_p}$ is the parameter vector, $f(\upsilon,p):\mathbb{R}^{n_0}\rightarrow\mathbb{R}$ is a continuously differentiable (and possibly nonconvex) function with a $L_f$-Lipschitz gradient, $\Upsilon\subseteq\mathbb{R}^{n_0}$ is a nonempty, closed set for computing projects, $F_1(\upsilon,p):\mathbb{R}^{n_0}\rightarrow\mathbb{R}^{n_1}$ is a smooth mapping with a Lipschitz-continuous Jacobian bounded on $U$, $\mathcal{C}\subseteq\mathbb{R}^{n_1}$ is a convex set to determine distances, and $F_2(\upsilon,p):\mathbb{R}^{n_0}\rightarrow\mathbb{R}^{n_2}$ is a function in which $||F_2(\upsilon,p)||^2$ is continuously differentiable with a Lipschitz-continuous gradient.

The constraints given by Eqs. \eqref{constraintla} and \eqref{constraintpen} are accounted in the optimization using the augmented Lagrangian method and the penalty method, respectively. Equality and inequality constraints are given by
\begin{subequations}
\begin{equation}
    H_{eq}(\upsilon,p)=0,
\end{equation}
\begin{equation}\label{ineqconstraint}
    H_{ineq}(\upsilon,p)\leq 0,
\end{equation}
\end{subequations}
and both can be included into Eqs. \eqref{constraintla} and \eqref{constraintpen}. By letting $F_1(\upsilon,p)=H_{eq}(\upsilon,p)$ in Eq. \eqref{constraintla}, the equality is incorporated by setting $\mathcal{C}=0$. The inequality constraint can be included in Eq. \eqref{constraintla} by letting $F_1(\upsilon,p)=H_{ineq}(\upsilon,p)$ and setting $\mathcal{C}=\{ \nu\in\mathbb{R}^{n_1}:\nu\leq0\}$. For Eq. \eqref{constraintpen}, the equality constraint can be found by letting $F_2(\upsilon,p)=H_{eq}(\upsilon,p)$ instead. The inequality constraint can be obtained  in Eq. \eqref{constraintpen} by letting $F_2(\upsilon,p)=\max\left[0, H_{ineq}(\upsilon,p)\right]$.

From the general parametric optimization problem given in Eq. \eqref{genoptprob}, a discrete-time single-shooting optimization problem can be formed as
\begin{subequations}\label{singleshooting}
\begin{equation}
    \min_{U_{0:N-1}}\sum_{k=0}^{N-1}l(F_k(U_{0:k},\mathbf{x}_0),\mathbf{u}_k)+l_N(F_N(U_{0:N-1},\mathbf{x}_0))
\end{equation}
\begin{align}
        \text{Subject} \hspace{5pt}\text{to}: &\mathbf{u}_k\in U_{0:N-1},\\
    &F_0(\mathbf{u}_0,\mathbf{x}_0)=\mathbf{x}_0,\\
    &F_{k+1}(U_{0:k},\mathbf{x}_0)=f(F_k(U_{0:k},\mathbf{x}_0),\mathbf{u}_k),\\
    &H_k(F_k(U_{0:k},\mathbf{x}_0),\mathbf{u}_k)\leq0\label{discreteconstraint},
\end{align}
\end{subequations}
where $F_k(U_{0:k},\mathbf{x}_0)$ is a sequence of functions to obtain $\mathbf{x}_k$, $f(\cdot,\mathbf{u}_k)$ is a nonlinear dynamics function given by Eq. \eqref{nonlinearequationdis}, and $H_k(F_k(U_{0:k},\mathbf{x}_0),\mathbf{u}_k)$ are the inequality constraints that can be formed as either Eq. \eqref{constraintla} or \eqref{constraintpen}. For a time-step $k$, the nonlinear MPC formulation can directly be derived from Eq. \eqref{singleshooting} by optimizing over a prediction horizon $N_p$ given by
\begin{subequations}\label{singleshootingmpc}
\begin{equation}
    \min_{U_{k:(k+N_p-1)}}\sum_{j=0}^{N_p-1}l(F_{k+j}(U_{0:(k+j)},\mathbf{x}_0),\mathbf{u}_{k+j})+l_{N_p}(F_{N_p}(U_{0:(N_p-1)},\mathbf{x}_0))
\end{equation}
\begin{align}
        \text{Subject} \hspace{5pt}\text{to}: &\mathbf{u}_{k+j}\in U_{k:(k+N_p-1)},\\
    &F_0(\mathbf{u}_0,\mathbf{x}_0)=\mathbf{x}_0,\\
    &F_{k+j+1}(U_{0:(k+j)},\mathbf{x}_0)=f(F_{k+j}(U_{0:(k+j)},\mathbf{x}_0),\mathbf{u}_{k+j}),\\
    &H_{k+j}(F_{k+j}(U_{0:(k+j)},\mathbf{x}_0),\mathbf{u}_{k+j})\leq0\label{discreteconstraint2}.
\end{align}
\end{subequations}
At each time-step $k$, an optimal control sequence of $U^{\star}_{k:(k+N_p-1)}=\left[\mathbf{u}^{\star}_{k},\mathbf{u}^{\star}_{k+1},\dots,\mathbf{u}^{\star}_{k+N_p-1} \right]$ is found, and the first control input, $\mathbf{u}^{\star}_{k}$, is applied to the robotic system (Eq. \eqref{nonlinearequationdis}). The prediction horizon, $N_p$, can be modified to to trade-off the computational performance of the optimization and the optimality of the solution with respect to the finite horizon $N$. With the optimization problem formed, the optimal control solutions can be obtained using the PANOC solver to guarantee real-time performance \cite{sopasakis2020open,stella2017simple}. In the next two sections, the obstacle avoidance constraints and the dynamical constraints are discussed, which are applied to either the LQR-RRT*, trajectory smoothing, or nonlinear MPC in specific forms.


\section{Collision Avoidance}
For on-orbit assembly, obstacle avoidance is an area to consider to prevent harm to either the on-orbit free-flyer or the assembled structure. Fortunately, collision avoidance can be implemented in all parts of the problem described in Fig. \ref{hierarchy}. A component of LQR-RRT* and trajectory smoothing via shortcutting include a collision checking element which checks whether the path between two nodes intersects with an obstacle. If there is such an intersection, the path between the two nodes is thrown out. The LQR planning algorithm applied after the smoothing does not guarantee collision avoidance, but the motion planning can be designed to follow the shortcutting trajectory to prevent obstacle collisions. Also as discussed previously, the LQR motion planning can be incorporated into the \textit{Straight Line Interpolation} step of the shortcutting algorithm to provide time-dependent trajectories with guaranteed collision avoidance. Lastly, collision avoidance can be implemented directly as a constraint for the nonlinear MPC problem. By solving the nonlinear MPC problem using a PANOC solver, control solutions can be found that follows a trajectory while avoiding obstacles.

For the on-orbit free-flyer, the goal is to avoid obstacle collisions with structural parts during its structural assembly. The obstacles themselves can be tracked by the on-orbit free-flyer or remotely by another spacecraft or a ground station. By knowing the states of the obstacles, the on-orbit free-flyer can approximate a keep-out zone using a 3D ellipsoid \cite{jewison2015model}.
\subsection{Ellipsoid Method}
A 3D ellipsoid constraint bounds an obstacle by an ellipsoidal keep-out zone. If the obstacle has no uncertainty in its size, an ellipsoid can be formed which encloses the obstacle's volume. If uncertainty in the obstacle's size exists, the ellipsoid that encloses the obstacles may include a safety factor. For estimation problems with assumed Gaussian noise, the corresponding covariance can be directly incorporated into the ellipsoid model to represent the uncertainty of the obstacle inside of it. The ellipsoid representation of obstacles can also be directly applied to higher fidelity applications. Instead of bounding a entire space structure with an ellipsoid, individual structural parts can be modeled with an ellipsoid to be avoided. This is useful for on-orbit assembly applications, since individual structural parts are manipulated in order to assemble larger structures. The ellipsoidal constraint for collision avoidance is given by
\begin{equation}\label{ellcon}
    \left( \mathbf{x}_{pos}-\mathbf{x}_{obs}\right)^T P_{obs}\left( \mathbf{x}_{pos}-\mathbf{x}_{obs}\right)\geq 1,
\end{equation}
where $\mathbf{x}_{obs}$ is the ellipsoid's centroid position, $P_{obs}$ is a positive definite shape matrix of the ellipsoid, and $\mathbf{x}_{pos}$ is the position of a point on the path in question. This constraint is nonlinear and nonconvex, but motion planning and control discussed in the previous section are able to handle such constraints. If $\mathbf{x}_{pos}$ lies outside the ellipsoid, the constraint in Eq. \eqref{ellcon} is met (returned true), and if $\mathbf{x}_{pos}$ lie inside the ellipsoid, the constraint in Eq. \eqref{ellcon} is not met (returned false). Each point on the trajectory must be evaluated against Eq. \eqref{ellcon} to determine whether any collisions occurred. For LQR-RRT* and trajectory smoothing using shortcutting, this is a simple for-loop. For nonlinear MPC, the constraint must be formed as Eq. \eqref{ineqconstraint} given by
\begin{equation}\label{ellconfull}
    H_{ineq}=\begin{bmatrix}
          1 \\
           \vdots \\
           1
         \end{bmatrix}_{N\times 1} -\text{diag}\{ \mathcal{X}^T P_{obs}\mathcal{X}\}\leq \begin{bmatrix}
          0 \\
           \vdots \\
           0
         \end{bmatrix}_{N\times 1}
\end{equation}
where $N$ is the time horizon of the trajectory, $\mathcal{X}$ is a stacked sequence of position states specified by $\mathcal{X}=\left[ \left(\mathbf{x}_{pos,0}-\mathbf{x}_{obs,0}\right),\left(\mathbf{x}_{pos,1}-\mathbf{x}_{obs,1}\right),\dots,\left(\mathbf{x}_{pos,N}-\mathbf{x}_{obs,N}\right)\right]\in\mathbb{R}^{3\times N}$, and $\text{diag}(\cdot)$ transforms the diagonals of the $N\times N$ matrix into a column vector. Note that $\mathbf{x}_{obs}$ can be dynamically moving, and thus, a time dependency is included in Eq. \eqref{ellconfull}. This provides the collision check or constraint for a single obstacle, but this constraint can be expanded to multiple obstacles by comparing the trajectory position state to every obstacle on the field. This is given by
\begin{equation}\label{ellconful2l}
    H_{ineq}=\begin{bmatrix}
          1 \\
           \vdots \\
           1
         \end{bmatrix}_{(N\cdot N_{obs})\times 1} -\begin{bmatrix}\text{diag}\{ \mathcal{X}^T P_{obs,1}\mathcal{X}\}\\
         \vdots \\
         \text{diag}\{ \mathcal{X}^T P_{obs,N_{obs}}\mathcal{X}\}
         \end{bmatrix}
         \leq \begin{bmatrix}
          0 \\
           \vdots \\
           0
         \end{bmatrix}_{(N\cdot N_{obs})\times 1}
\end{equation}
where $N_{obs}$ is the number of obstacles on the field. Thus, obstacle avoidance using the ellipsoidal method can be applied to LQR-RRT*, trajectory smoothing using shortcutting, and nonlinear MPC using PANOC optimization. 

Since, Eq. \eqref{ellcon} is a nonconvex problem, considerations must be discussed with ill-posed planning problems. If the velocity of the robot is considerably fast, the time update between states in a discrete system can jump a great deal if the time interval is not fine enough. If this distance between two consecutive states is larger than the smallest characteristic length of the ellipsoid, then optimization solutions can be found which meet the constraints, but the robot technically collides with the obstacle. Great care must be taken to prevent these ill-posed problems, and one method to mitigate them is by increasing the resolution of the sample time and interpolating the trajectory in the obstacle constraint comparison \cite{jewison2017guidance}. Therefore, collision-free trajectories can be obtained from the optimization solutions.
\section{Robotic Free-Flyer Dynamics}
To show the viability of the motion planning and control of an on-orbit free-flyer for assembly, a multi-rigid body system is used to describe the dynamics. The dynamics are used directly in motion planning and control of the spacecraft by computing the corresponding linearization and discretization discussed in the Appendix.
\subsection{Multi-Rigid Body Dynamics}
For the formulation of the on-orbit free-flyer dynamics, it is assumed that the spacecraft consists of multiple rigid body parts including a base and a manipulator. Also, it is assumed that the effects due to relative motion are negligible since the structural parts are within a close vicinity of the spacecraft. 

The states for the multi-body, on-orbit free-flyer consists of
\begin{equation}
    \mathbf{x}_{base}(t)=\begin{bmatrix}
          \mathbf{r}_{base}(t) \\
          \boldsymbol{\theta}_{base}(t)
         \end{bmatrix},\hspace{12pt}
         \mathbf{q}_{m}(t)=\begin{bmatrix}
          \mathbf{q}_{1}(t) \\
          \vdots\\
          \mathbf{q}_{N_m}(t)
         \end{bmatrix},
\end{equation}
where $\mathbf{r}_{base}(t)$ and $\boldsymbol{\theta}_{base}(t)$ are the Cartesian position and attitude of the base spacecraft and $\mathbf{q}_m(t)$ are the joint angular positions of the manipulator at a time $t$. The manipulator state, $\mathbf{q}_m(t)$, is generalized with $N_m$ number of joints. The multiple rigid body states can be combined into one state specified by
\begin{equation}
    \mathbf{x}(t)=\begin{bmatrix}
          \mathbf{x}_{base}(t) \\
          \mathbf{q}_{m}(t)
         \end{bmatrix}.
\end{equation}
This multi-body state description can be propagated through time in a compact nonlinear form
\begin{equation}\label{roboteom}
    G(\mathbf{x}(t))\ddot{\mathbf{x}}(t)+D(\mathbf{x}(t),\dot{\mathbf{x}}(t))\dot{\mathbf{x}}(t)=\boldsymbol{\tau}(t),
\end{equation}
where $G(\mathbf{x}(t))\in\mathbb{R}^{(6+N_m)\times(6+N_m)}$ is a positive-semidefinite, symmetric, generalized inertia matrix, $D(\mathbf{x}(t),\dot{\mathbf{x}}(t))\in\mathbb{R}^{(6+N_m)\times(6+N_m)}$ is a generalized convective inertia matrix which contains the nonlinear centrifugal and Coriolis expressions, and $\boldsymbol{\tau}(t)$ are the generalized torques and forces applied on the state-space \cite{dubowsky1993kinematics}. The generalized inertia and convective inertia matrices, $G(\mathbf{x}(t))$ and $D(\mathbf{x}(t),\dot{\mathbf{x}}(t))$, can be computed directly using efficient recursions obtained through the Newton-Euler approach and Decoupled Natural Orthogonal Complement Matrices (DeNOC) \cite{saha1999dynamics,shah2012modular}. These methods provide efficient algorithms to compute the forward and inverse kinematics necessary to derive $G(\mathbf{x}(t))$ and $D(\mathbf{x}(t),\dot{\mathbf{x}}(t))$. Specifically, the Spacecraft Robotics Toolkit (SPART) models the multi-robotic system's equations of motion from Eq. \eqref{roboteom} using this approach  \cite{virgili2016spacecraft}. From a URDF (Unified Robot Description Format) file of the on-orbit free-flyer, SPART is used to obtain the dynamics for LQR-RRT*, trajectory smoothing, and nonlinear MPC using the PANOC solver. 

\section{Simulation Results}
For this work, the Astrobee on-orbit free-flyer, with a simplified 2-Degree of Freedom (DOF) manipulator, is tasked for on-orbit assembly of space structures. This includes the motion planning to and from a 3D printer (in which parts are manufactured) and the area which the space structure is built. One top of the simplified manipulator, it is also assumed that the actuation is not constrained to any limits, and during the manipulation, the mass properties do not change between the combined Astrobee and structural part. 

 \begin{table}[htbp]
   \caption{Astrobee: Inertial Properties} \label{tab:AIP}
   \small 
   \centering 
   \begin{tabular}{lcccc} 
   \toprule[\heavyrulewidth]\toprule[\heavyrulewidth]
   \textbf{Properties} & \textbf{Mass [kg]}&\multicolumn{3}{c}{\textbf{Moment of Inertia [kg m$^\text{2}$]}} \\ 
    & & \textbf{I$_{\text{xx}}$} &\textbf{I$_{\text{yy}}$}&\textbf{I$_{\text{zz}}$}\\
   \midrule
   Base   & 7.0&0.11&0.11&0.11 \\
   Arm 1 & 1.0 &0.05&0.05&0.05\\
   Arm 2 & 1.0 &0.05&0.05&0.05\\
   End-Effector & 4.0 &0 &0&0\\

   \bottomrule[\heavyrulewidth] 
   \end{tabular}
\end{table}

The base Astrobee has a full (6-DOF) capability inside the state-space environment and the manipulator has an additional 2-DOF arm ($N_m=2$) from actuation of the joints \cite{alsup2018robotic}. The mass properties for a simulated Astrobee used in this problem are given in Table \ref{tab:AIP} \cite{alsup2018robotic}. Note that the properties are modified for this problem (e.g. the end-effector is assumed to have a mass of 4kg as consideration for any structural parts attached). From these properties, A URDF robotic description file of Astrobee was designed to obtain the multi-body dynamics necessary for motion planning and control using the hierarchy in Fig. \ref{hierarchy}. The results obtained through this hierarchy are discussed next.

\begin{figure}[!htb]
\begin{centering}
    \subfigure[Base Translational States]{
\includegraphics[keepaspectratio,trim={2.35cm 1.75cm 3.1cm 1.75cm},clip,width=.48\textwidth]{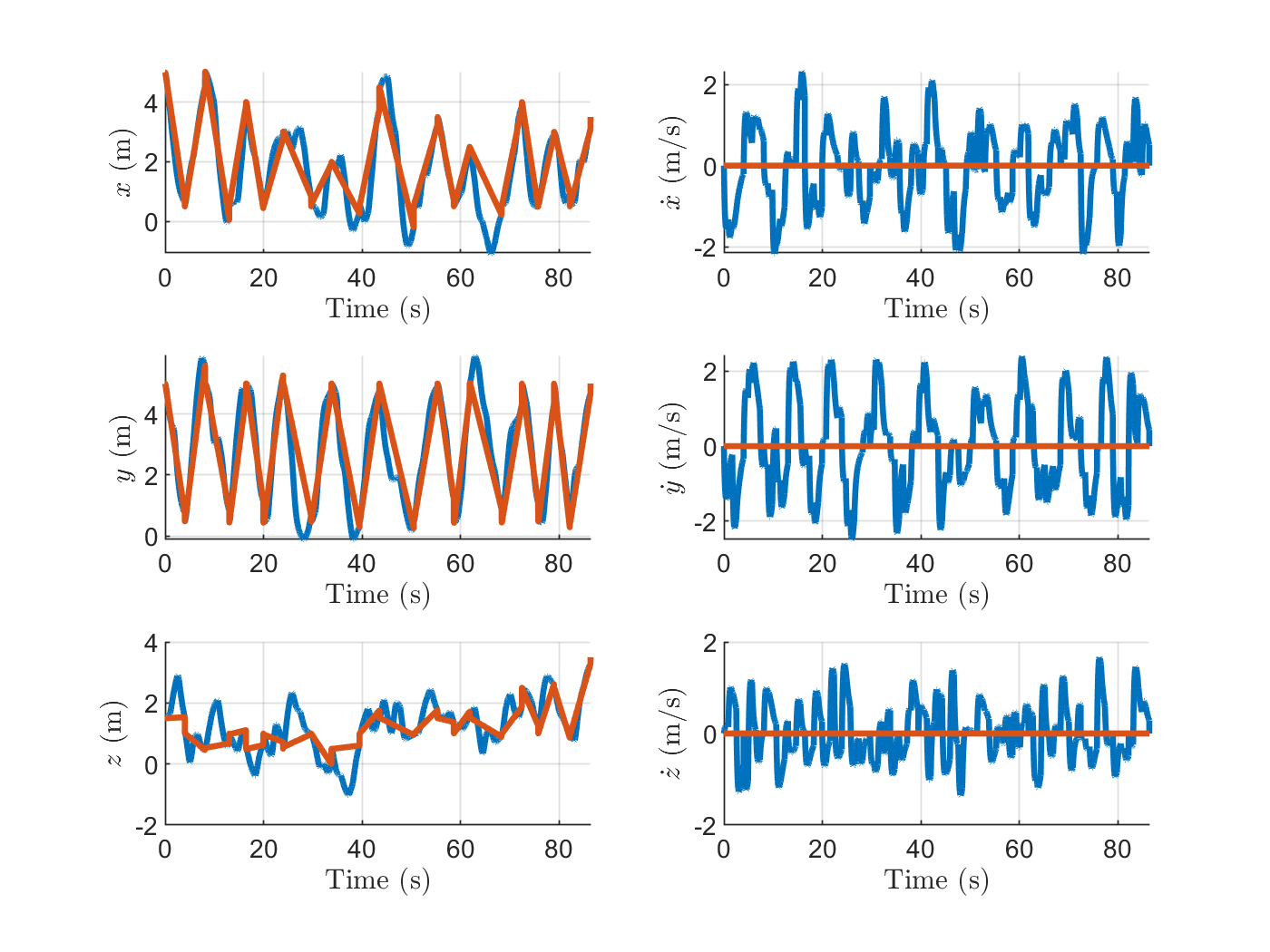}
      \label{fig:baseposrrt}}
\subfigure[Base Attitude]{
\includegraphics[keepaspectratio,trim={2.35cm 1.75cm 3.1cm 1.75cm},clip,width=.48\textwidth]{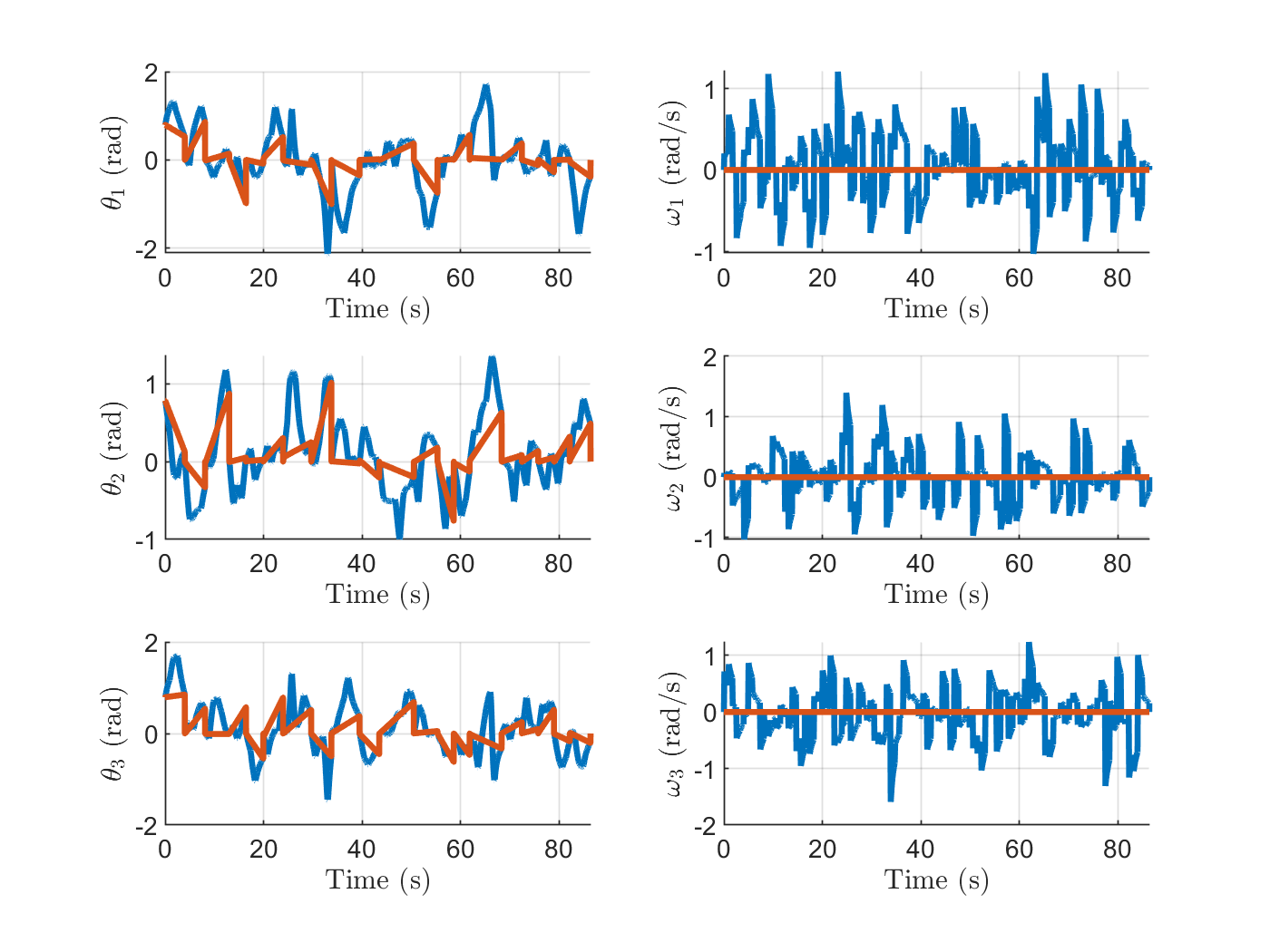}
    \label{fig:baseattrrt}} 
        \caption{Translational and attitude states for the base Astrobee using LQR-RRT* (blue) and trajectory smoothing (red).}\label{fig:lqrrrtresult}
\end{centering}
\end{figure}

\subsection{LQR-RRT*, Trajectory Smoothing, and LQR}
In order to assemble structures from 3D printed parts on-orbit, a simulation using LQR-RRT* was developed. For this scenario in Fig. \ref{hierarchy}, the goal is to find an initial path from Astrobee's current position to the 3D printer or to the desired location for construction. The trajectories obtained from LQR-RRT* are shown in blue in Figure \ref{fig:lqrrrtresult}. The trajectories were found to move between the 3D printer and the desired location for construction while avoiding ellipsoidal obstacles. At specific points along the path, the trajectory is jerky and unnatural since the algorithm is ran until an initial trajectory, which completes the path between Astrobee and the desired state, is found. If the LQR-RRT* algorithm continues sampling the space through time, the trajectories obtained would reach asymptotic optimality with respect to the cost function, but for this work, computational efficiency is necessary for Astrobee to perform on-orbit assembly. 

\begin{figure}[!htb]
\begin{centering}
    \subfigure[Base Translational States]{
\includegraphics[keepaspectratio,trim={2.35cm 1.75cm 3.1cm 1.75cm},clip,width=.48\textwidth]{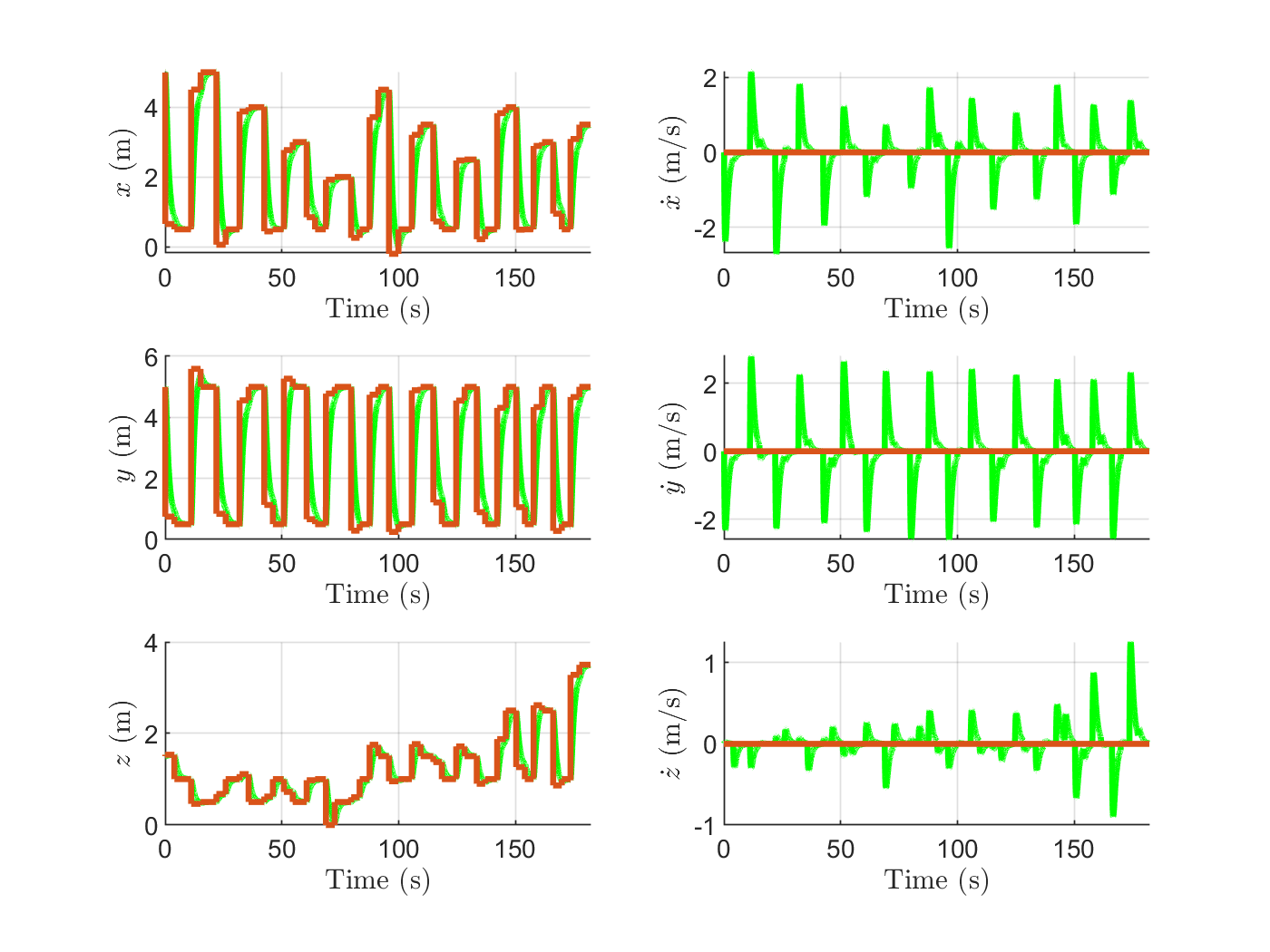}
      \label{fig:baseposlqr}}
\subfigure[Base Attitude]{
\includegraphics[keepaspectratio,trim={2.35cm 1.75cm 3.1cm 1.75cm},clip,width=.48\textwidth]{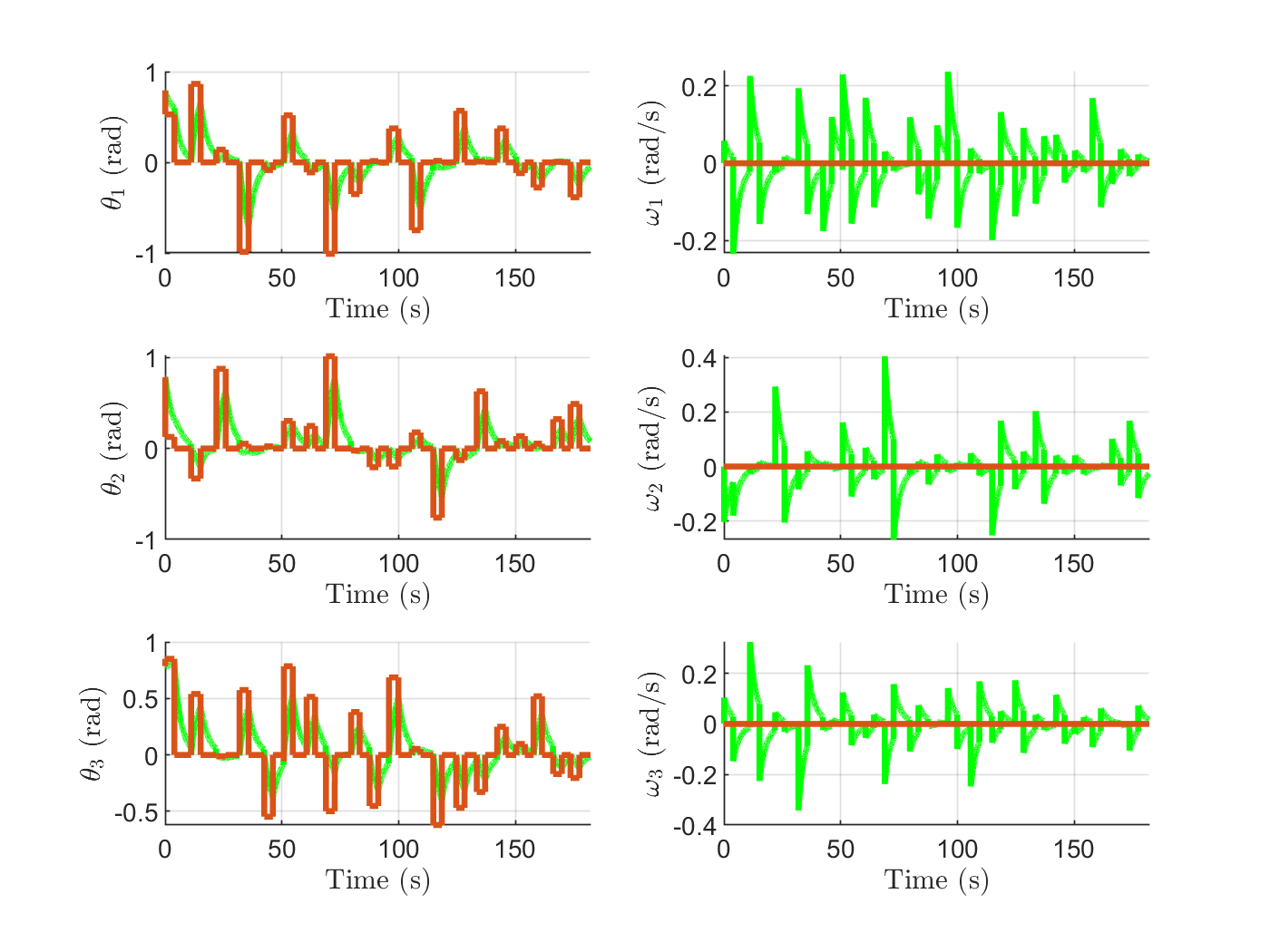}
    \label{fig:baseattlqr}} 
        \caption{Translational and attitude states for the base Astrobee using LQR (green) and the reference smoothing trajectory(red).}\label{fig:lqr}
\end{centering}
\end{figure}

To mitigate the randomness of sampling in LQR-RRT*, trajectory smoothing by shortcutting is implemented as shown in red within Figure \ref{fig:lqrrrtresult}. Specifically, smoother, collision-free paths (in position and orientation) are found for Astrobee, and mitigation of the randomness of the sampling from LQR-RRT* can be observed. As a note from the figure, each point of the shortcutting trajectory is time independent. This trajectory was projected onto the LQR-RRT* trajectory to show the differences between the two trajectories. 

\begin{figure}[!htb]
\begin{centering}
    \subfigure[Base Translational States]{
\includegraphics[keepaspectratio,trim={2.35cm 1.75cm 3.1cm 1.75cm},clip,width=.48\textwidth]{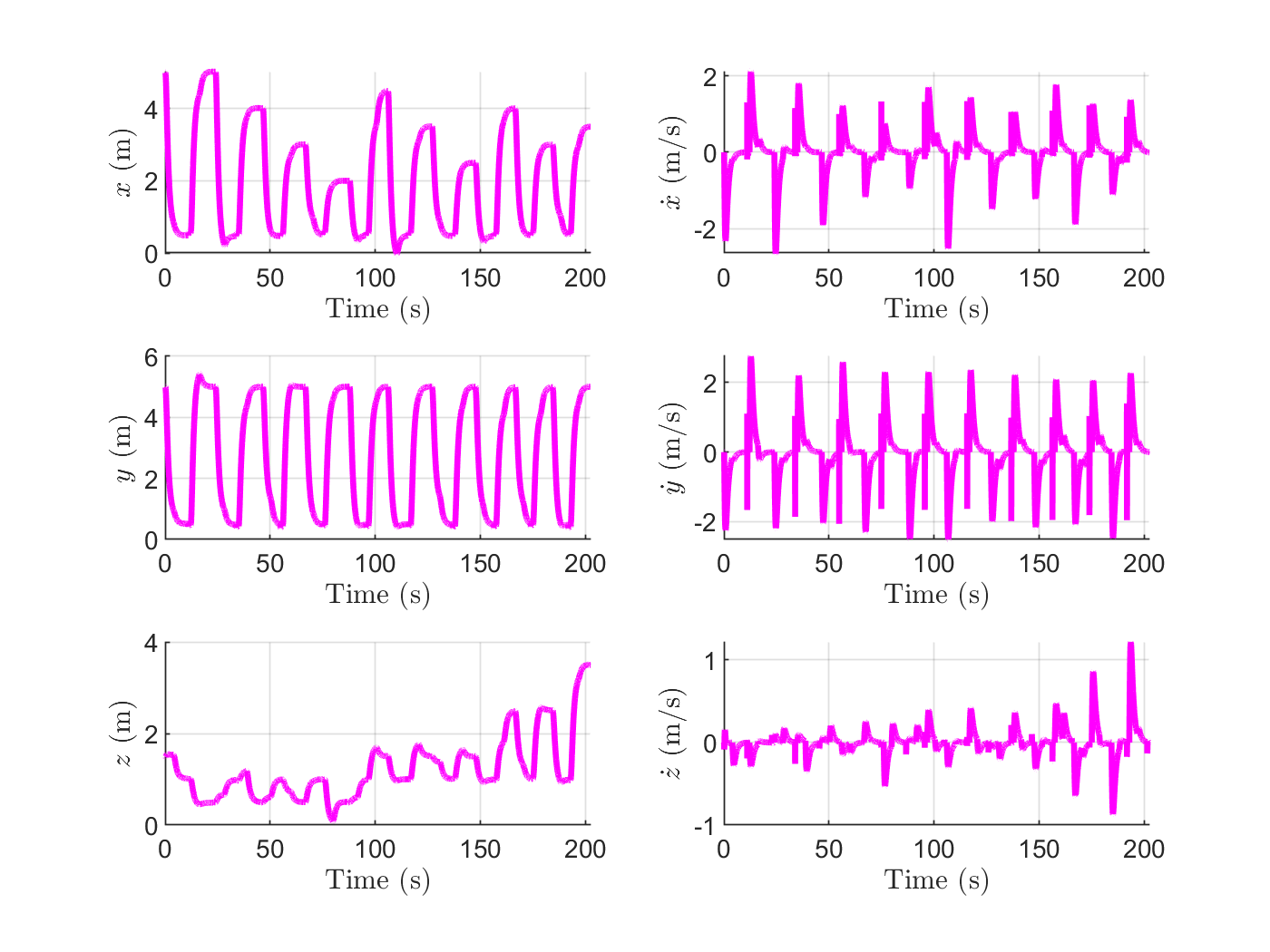}
      \label{fig:baseposopt}}
\subfigure[Base Attitude]{
\includegraphics[keepaspectratio,trim={2.35cm 1.75cm 3.1cm 1.75cm},clip,width=.48\textwidth]{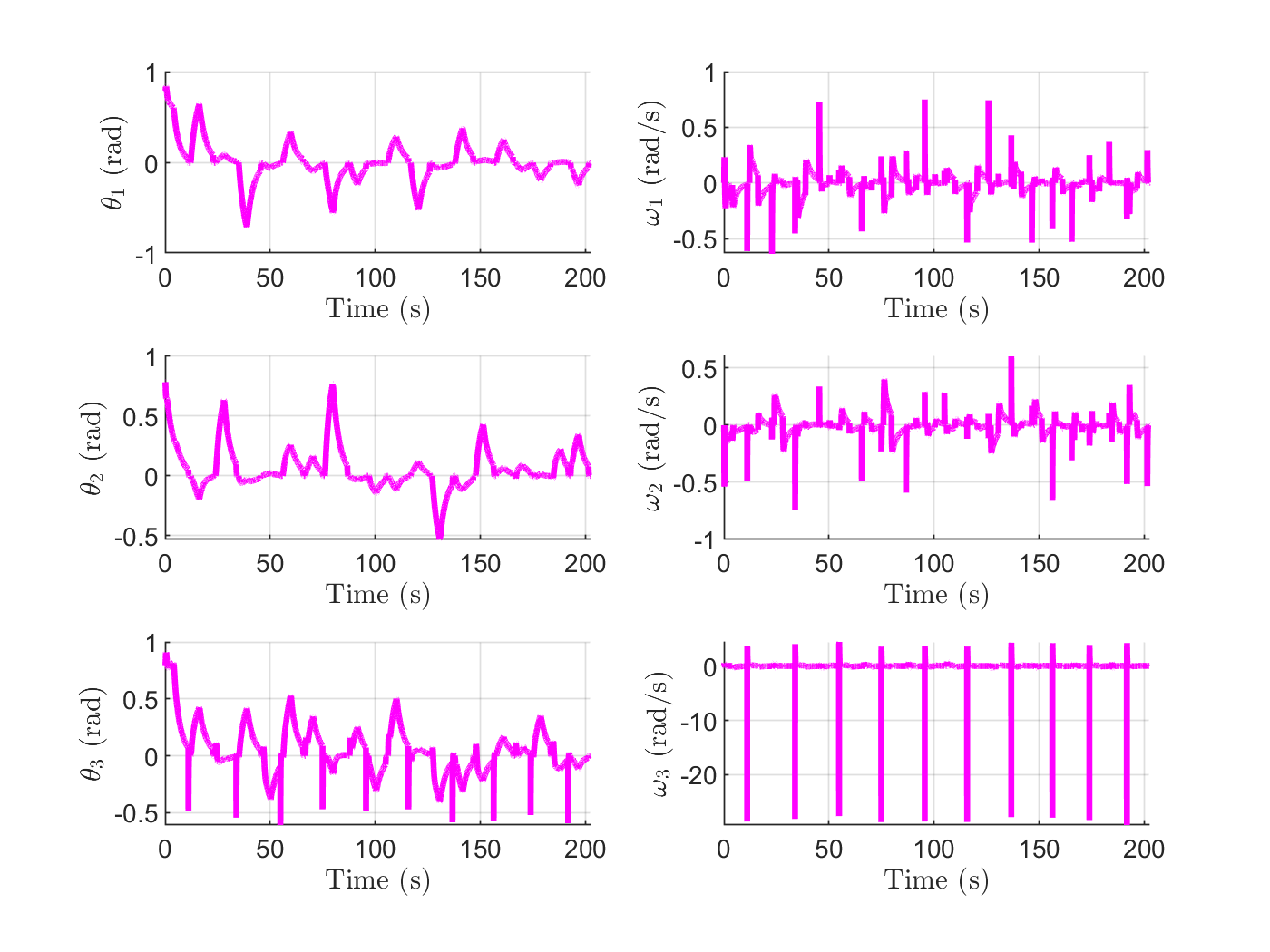}
    \label{fig:baseattopt}} 
\subfigure[Manipulator States]{
\includegraphics[keepaspectratio,trim={2.1cm .90cm 3.0cm 1.75cm},clip,width=.48\textwidth]{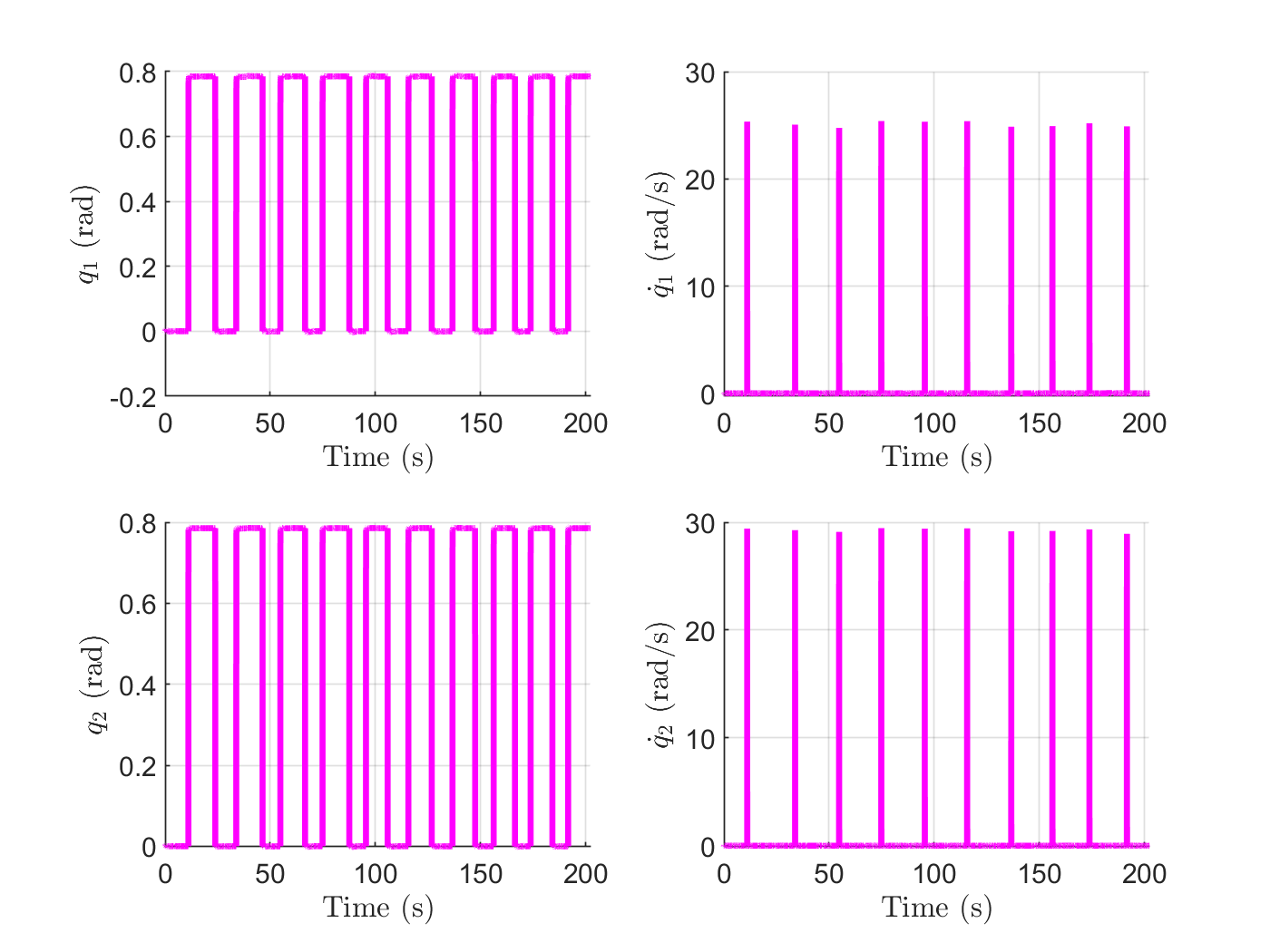}
    \label{fig:manattopt}} 
        \caption{Translational and attitude states for the base Astrobee and angular position states for the manipulator using nonlinear MPC.}\label{fig:nmpc}
\end{centering}
\end{figure}

Since the smoothed trajectory is time-independent, the smoothed trajectory was recomputed using LQR which appears in Figure \ref{fig:lqr} where the green and red lines are the LQR and shortcut trajectories, respectively. In this case, the smoothed trajectory is some desired state the Astrobee must achieve by LQR tracking. From the figure, the response reaches the desired state with a settling time of $3.40$s. By using LQR-RRT*, trajectory smoothing, and LQR to obtain a on-orbit Astrobee trajectory, MPC through PANOC can be applied to control the spacecraft along this trajectory as well as manipulate parts for on-orbit assembly.

\begin{figure}[!htb]
\begin{centering}
    \subfigure[Motion to first part]{
\includegraphics[keepaspectratio,trim={0cm .00cm 0cm .0cm},clip,width=.48\textwidth]{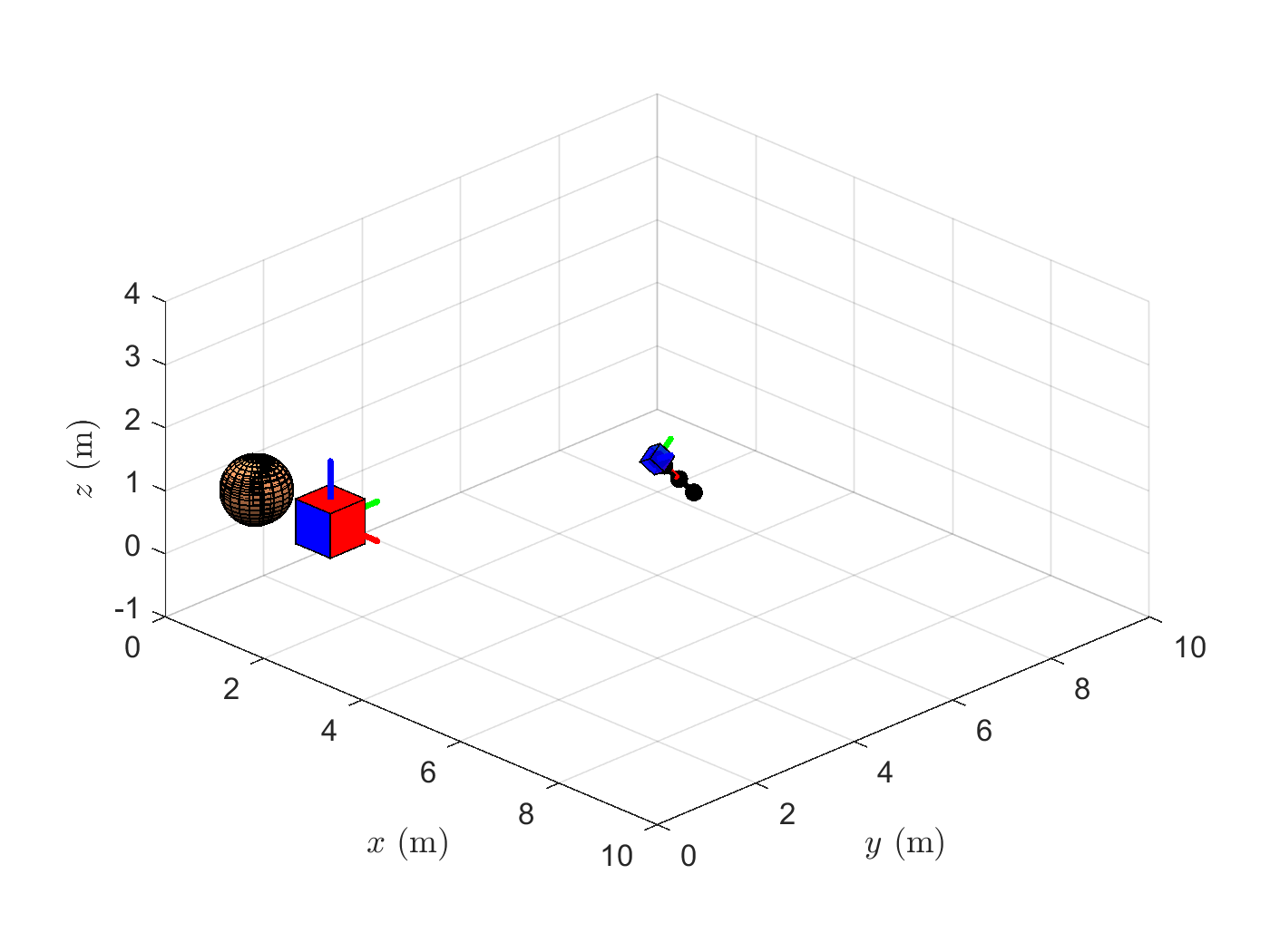}
      \label{fig:aa1}}
\subfigure[Manipulation of first part]{
\includegraphics[keepaspectratio,trim={0cm .00cm 0cm .0cm},clip,width=.48\textwidth]{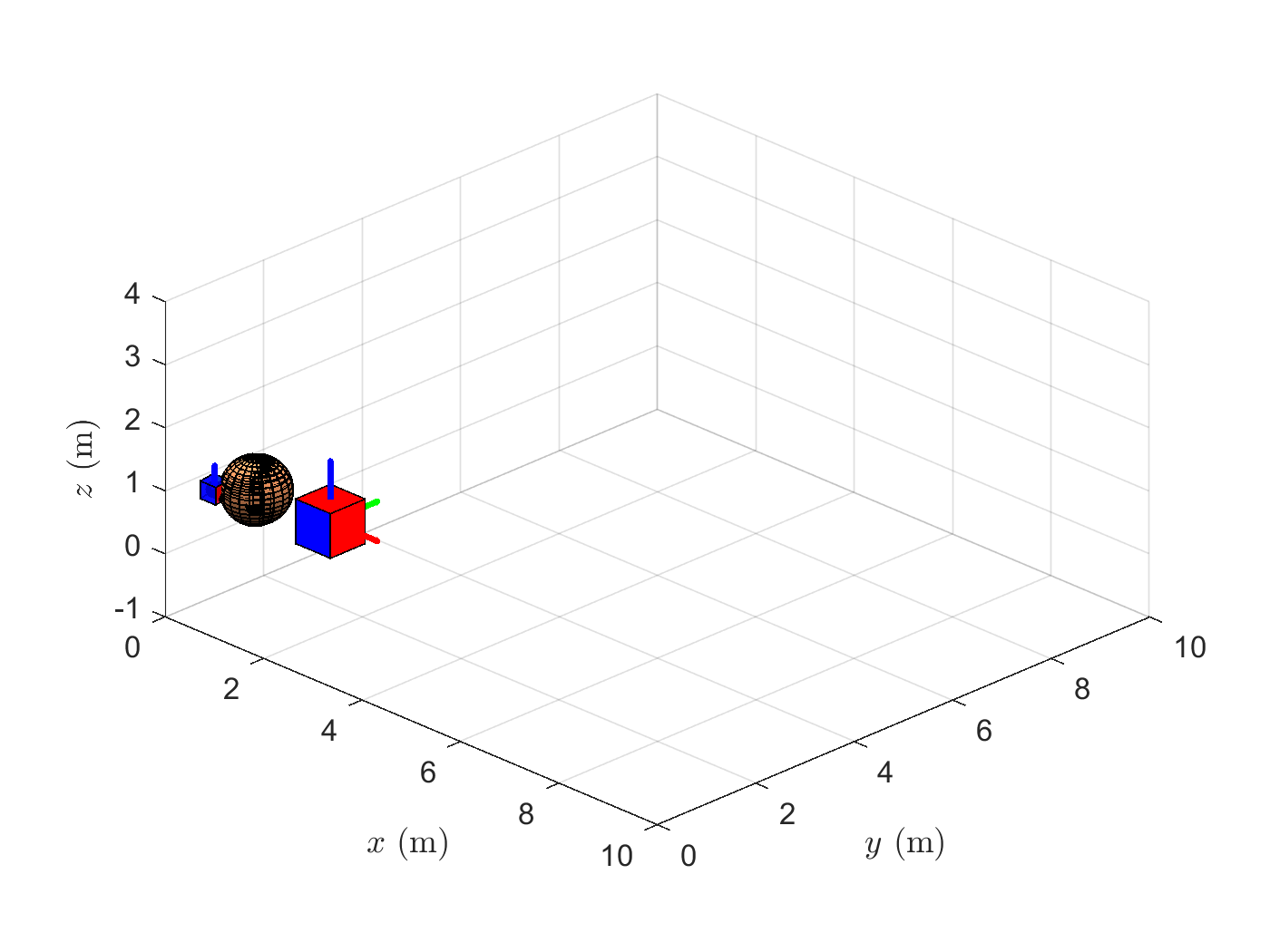}
    \label{fig:aa2}} 
\subfigure[Placement of part and motion back to 3D printer]{
\includegraphics[keepaspectratio,trim={0cm .00cm 0cm .0cm},clip,width=.48\textwidth]{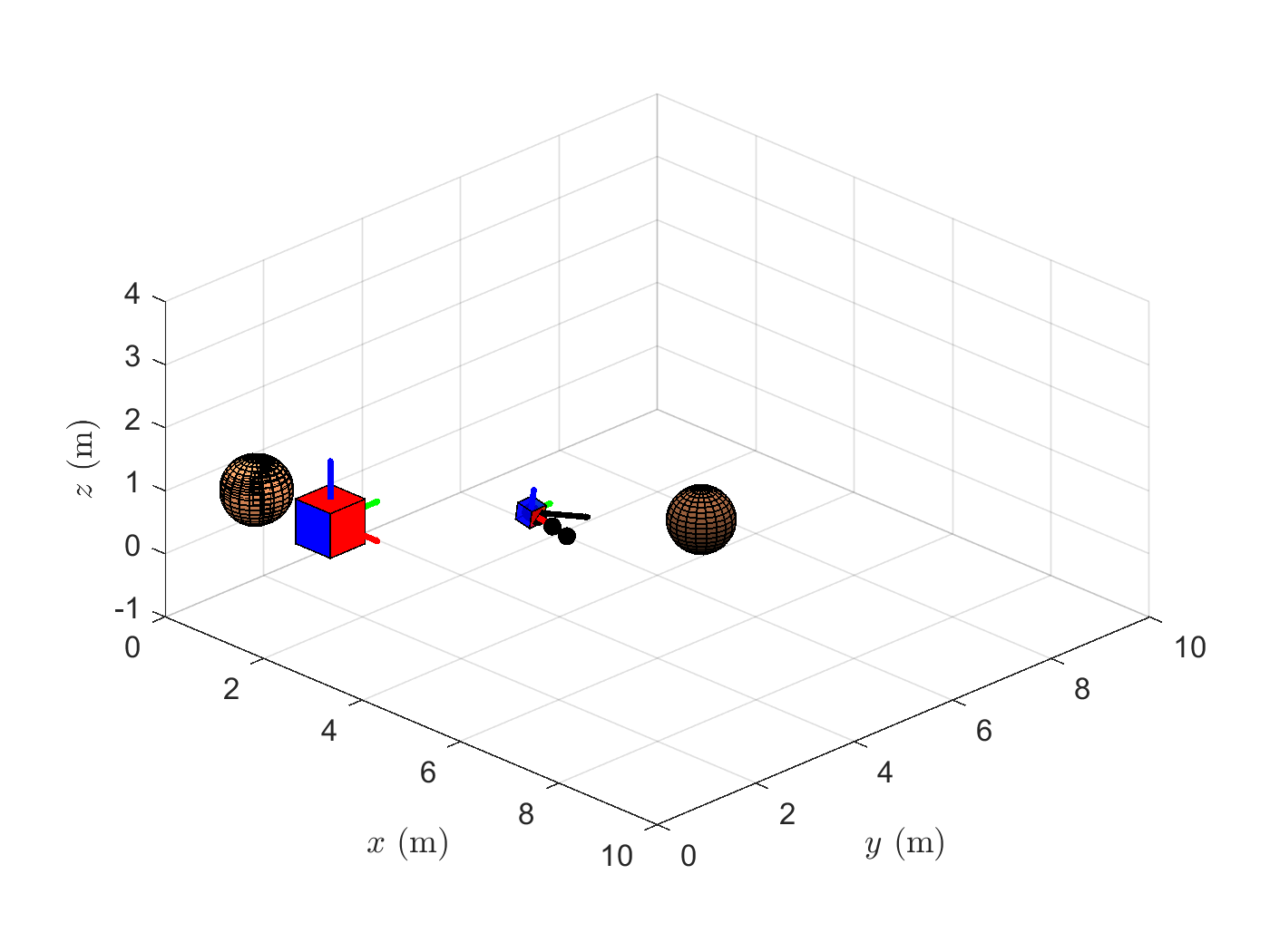}
      \label{fig:aa3}} 
\subfigure[Formation of bottom layer of structure]{
\includegraphics[keepaspectratio,trim={0cm .00cm 0cm .0cm},clip,width=.48\textwidth]{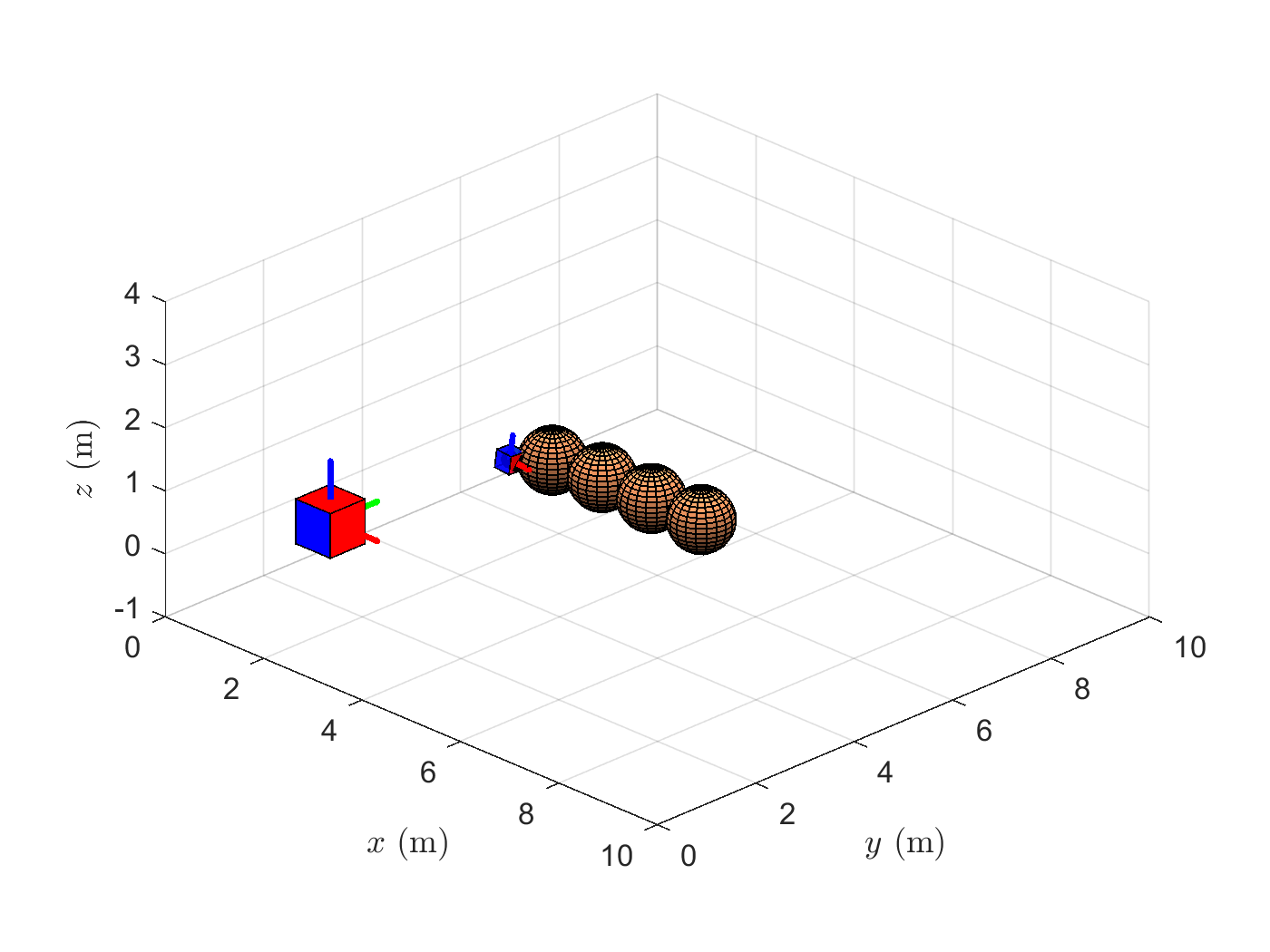}
      \label{fig:aa4}} 
\subfigure[Formation of second layer of structure]{
\includegraphics[keepaspectratio,trim={0cm .00cm 0cm .0cm},clip,width=.48\textwidth]{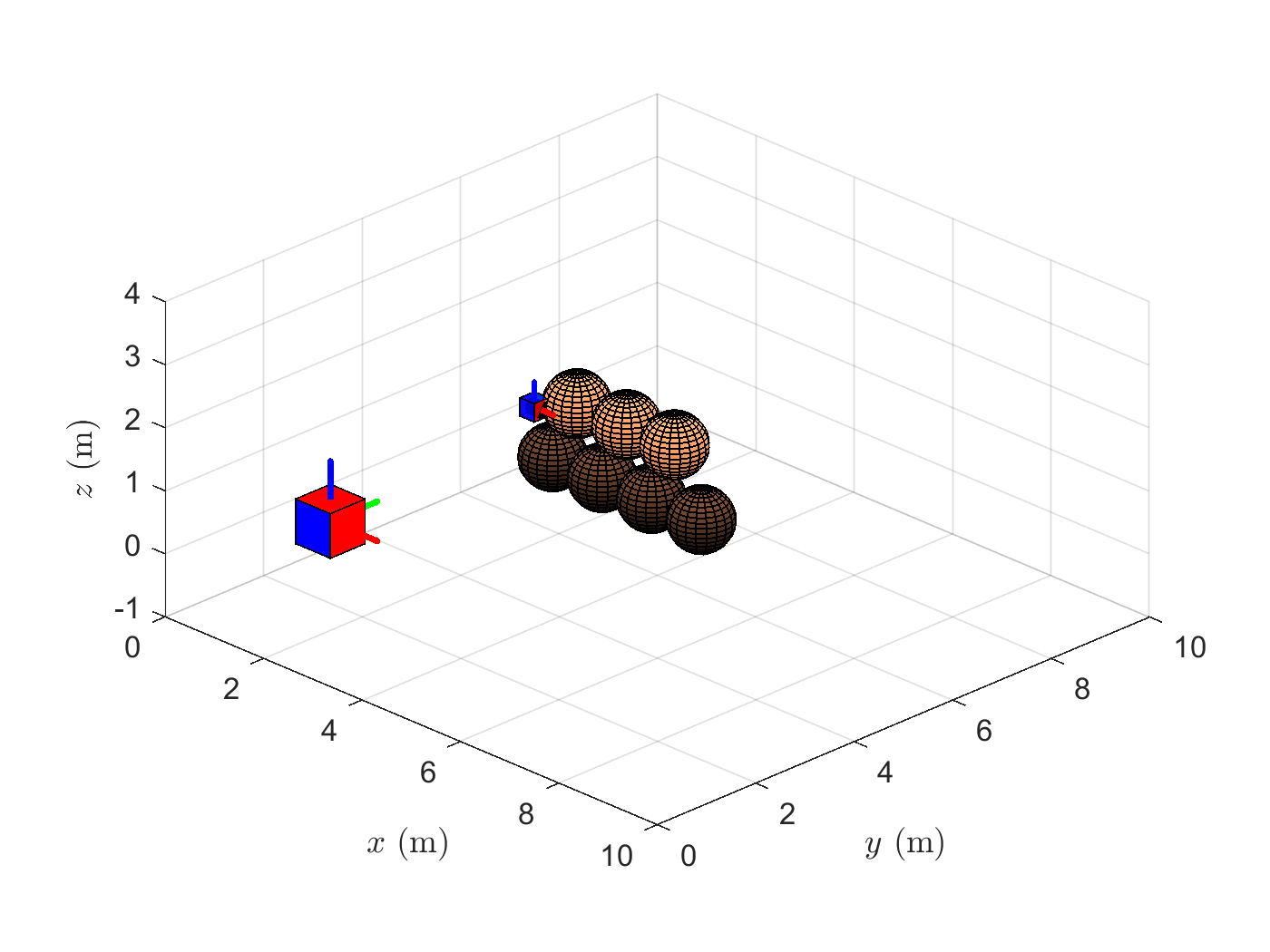}
      \label{fig:aa5}} 
\subfigure[The finished structure]{
\includegraphics[keepaspectratio,trim={0cm .00cm 0cm .0cm},clip,width=.48\textwidth]{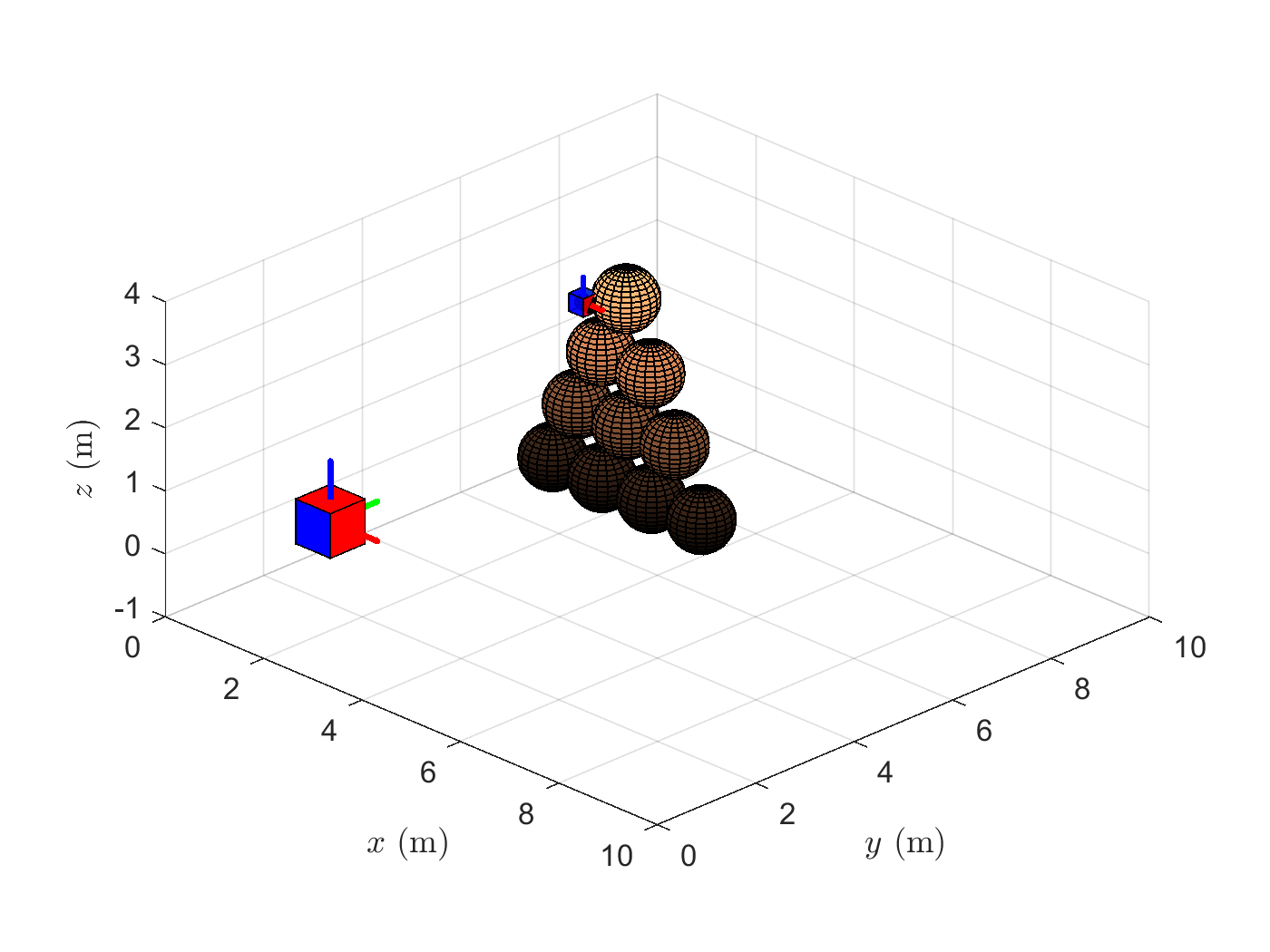}
      \label{fig:aa6}} 
        \caption{Trajectory snapshots of the on-orbit assembly process using Astrobee}\label{fig:snapshot}
\end{centering}
\end{figure}

\subsection{Nonlinear MPC using PANOC}
The trajectory obtained from LQR-RRT*, trajectory smoothing, and LQR was applied directly to the states of the base Astrobee. No trajectories were formed for the Astrobee manipulator during motion planning. Instead, the manipulator was commanded directly to a desired state using Nonlinear MPC. This is in addition to controlling the base Astrobee through the planned trajectory. The time-history response for the base and manipulator states of the Astrobee free-flyer during the on-orbit assembly is shown in Figure \ref{fig:nmpc}. From the figure, the manipulator was commanded to $\frac{\pi}{4}$ rad or $0$ rad to grasp the structural part or to retract during motion. The nonlinear MPC solver used a prediction horizon, $N_p$, of 10 time-steps into the future with an average of $30$ms computational time to solve for the control input, thus providing real-time capabilities.  Trajectory snapshots for the on-orbit assembly process is shown in Figure \ref{fig:snapshot}. The goal for Astrobee is to construct a pyramid-like structure with 10 ellipsoids constructed from a 3D printer (cube). Figure \ref{fig:aa1} shows Astrobee moving toward the first structural part. In Figure \ref{fig:aa2}, Astrobee grasps the first part and starts its motion to the assembly area. Next in Figure \ref{fig:aa3}, Astrobee places the first part in its desired area, and Astrobee moves onward to the second part. Figure \ref{fig:aa4} shows the bottom of the structure formed. Figure \ref{fig:aa5} shows the second layer of the structure formed. Figure \ref{fig:aa6} shows the finished pyramid-like structure formed from the ellipsoid structural part. When computing collision avoidance for the 3D printer and the structural parts, the $P_{obs}$ in Eq. \eqref{ellcon} was designed with a safety factor due to the size of the Astrobee from its centroid and trajectories obtained from LQR. Thus, control of collision-free trajectories for the on-orbit assembly system was obtained.

\section{Conclusion}
The objective of the paper is to formulate the motion planning and control of a robotic free-flyer (Astrobee) for on-orbit assembly of space structures through the extension of LQR-RRT*, trajectory smoothing, obstacle avoidance, and nonlinear MPC using PANOC. By setting up a problem in which parts manufactured by a 3D printer are described by an ellipsoid, collision-free trajectories are found for the on-orbit free-flyer. Then, real-time nonlinear MPC solutions were determined to maintain a collision-free optimal trajectory following the 8-DOF free-flyer and manipulator dynamics. This provides the on-orbit free-flyer the ability for on-orbit manipulation and the rapid consideration for the construction of space structures in simulation. For future work, the goal is to extend this work to physical experiments with the Astrobee robot on the ISS, identify changes in moment of inertia during manipulation, characterize contact during the manipulation, and improve trajectory smoothing by finding time-dependent solutions.

\section*{Acknowledgment}
This research was supported by an appointment to the Intelligence Community Postdoctoral Research Fellowship Program at Massachusetts Institute of Technology, administered by Oak Ridge Institute for Science and Education through an interagency agreement between the U.S. Department of Energy and the Office of the Director of National Intelligence. The authors wish to acknowledge useful conversations related to motion planning and optimization with Keenan Albee, a Ph.D. student, at the Massachusetts Institute of Technology.
\appendix
\section*{Appendix: Linearization and Discretization}
\subsection*{Linearization}
The dynamics for a free-flyer robot is described by the nonlinear equations of motion given by Eq. \eqref{fxu}. In order to apply LQR-RRT*, the equations of motion must be linearized and discretized. By taking a first-order Taylor series expansion of Eq. \eqref{fxu} around an operating point for linearization, $\bar{x}(t)$ and $\bar{u}(t)$, a linear time-varying (LTV) system can be found which approximates the nonlinear equation given by
\begin{equation}
\resizebox{1 \hsize}{!}{$
    \dot{\mathbf{x}}(t)\approx f(\mathbf{x}(t),\mathbf{u}(t))\vert_{\bar{\mathbf{x}}(t),\bar{\mathbf{u}}(t)}+\left.\frac{\partial f(\mathbf{x}(t),\mathbf{u}(t))}{\partial\mathbf{x}(t)}\right\vert_{\bar{\mathbf{x}}(t),\bar{\mathbf{u}}(t)}\left( \mathbf{x}(t)-\bar{\mathbf{x}}(t)\right)+\left.\frac{\partial f(\mathbf{x}(t),\mathbf{u}(t))}{\partial\mathbf{u}(t)}\right\vert_{\bar{\mathbf{x}}(t),\bar{\mathbf{u}}(t)}\left( \mathbf{u}(t)-\bar{\mathbf{u}}(t)\right).$}
\end{equation}
This equation is simplified to a linear state-space model using perturbations from the operating point given by
\begin{equation}\label{partiallinsys}
    \delta\dot{\mathbf{x}}(t)=\tilde{A}(t)\delta\mathbf{x}(t)+\tilde{B}(t)\delta\mathbf{u}(t),
\end{equation}
where $ \delta\dot{\mathbf{x}}(t)=\dot{\mathbf{x}}(t)-f(\mathbf{x}(t),\mathbf{u}(t))\vert_{\bar{\mathbf{x}}(t),\bar{\mathbf{u}}(t)}$, $\delta\mathbf{x}(t)=\mathbf{x}(t)-\bar{\mathbf{x}}(t)$, and $\delta\mathbf{u}(t)=\mathbf{u}(t)-\bar{\mathbf{u}}(t)$. Both $\tilde{A}(t)$ and $\tilde{B}(t)$ are time-dependent Jacobians of the nonlinear function $f(\mathbf{x}(t),\mathbf{u}(t))$ with respect to $\bar{\mathbf{x}}(t)$ and $\bar{\mathbf{u}}(t)$.
\subsection*{Discretization}
Either the full nonlinear dynamics given by Eq. \eqref{fxu} or the approximate linear dynamics given by Eq. \eqref{partiallinsys} can be discretized using a fourth-order Runge-Kutta method with a zero-order hold on the control input $\mathbf{u}_k$ \cite{van1978computing}. A fourth-order Runge-Kutta integration for Eq. \eqref{partiallinsys} is given by
\begin{equation}\label{4zoh}
    \delta\mathbf{x}_{k+1}=\delta\mathbf{x}_{k}+\frac{1}{6}h(k_1+2k_2+2k_3+k_4).
\end{equation}
Note that the variable $k$ is the discretized time-step and $h$ is the step size for the system. With the zero-order hold on a control input $\mathbf{u}_k$, the $k_1$, $k_2$, $k_3$, and $k_4$ are given by
\begin{equation}\label{kkkk}
    \begin{split}
        k_1&=\tilde{A}_k\delta\mathbf{x}_k+\tilde{B}_k\delta\mathbf{u}_k\\
        k_2&=\tilde{A}_k\left(\delta\mathbf{x}_k+k_1/2\right)+\tilde{B}_k\delta\mathbf{u}_k\\
        k_3&= \tilde{A}_k\left(\delta\mathbf{x}_k+k_2/2\right)+\tilde{B}_k\delta\mathbf{u}_k\\
        k_4&=\tilde{A}_k\left(\delta\mathbf{x}_k+k_3\right)+\tilde{B}_k\delta\mathbf{u}_k.
    \end{split}
\end{equation}
By substituting Eq. \eqref{kkkk} into Eq. \eqref{4zoh}, the discretized system from a fourth-order Runge-Kutta with a zero-order hold on control is
\begin{equation}\label{discretization}
    \begin{gathered}
    \delta\mathbf{x}_{k+1}=\left( I+h\tilde{A}_k+\frac{h^2}{2!}\tilde{A}_k^2+\frac{h^3}{3!}\tilde{A}_k^3+\frac{h^4}{4!}\tilde{A}_k^4\right)\delta\mathbf{x}_{k}+\left( h+\frac{h^2}{2!}\tilde{A}_k+\frac{h^3}{3!}\tilde{A}_k^2+\frac{h^4}{4!}\tilde{A}_k^3\right)\tilde{B}_k\delta\mathbf{u}_k,
    \end{gathered}
\end{equation}
which follows the form of Eq. \eqref{axbu}. This discrete, linear approximation for Eq. \eqref{fxu} is used to produce trajectories using LQR-RRT*. Similarly, the a discrete, nonlinear form of Eq. \eqref{fxu} can be obtained for nonlinear MPC using the same fourth-order Runge-Kutta method which results in Eq. \eqref{nonlinearequationdis}.
\bibliographystyle{AAS_publication}   
\bibliography{references}   

\end{document}